\documentclass{article}
\usepackage{iclr2027_conference,times}

\usepackage[utf8]{inputenc}
\usepackage[T1]{fontenc}
\usepackage{url}
\usepackage{booktabs}
\usepackage{multirow}
\usepackage{wrapfig}
\usepackage{graphicx}
\usepackage{placeins}
\usepackage{float}
\usepackage{flafter}
\usepackage{needspace}
\usepackage{titletoc}
\usepackage{amsmath}
\usepackage{amsfonts}
\usepackage{amssymb}
\usepackage{pifont}
\newcommand{\cmark}{\ding{51}}
\newcommand{\xmark}{\ding{55}}
\usepackage{nicefrac}
\usepackage{microtype}
\usepackage[table]{xcolor}
\definecolor{best_result}{rgb}{0.69, 0.81, 0.94}
\definecolor{second_result}{rgb}{0.92, 0.95, 0.82}
\usepackage{tikz}
\usetikzlibrary{arrows.meta}
\usepackage{hyperref}
\usepackage[all]{hypcap}
\usepackage{cleveref}
\hypersetup{colorlinks=true,pdftitle={Forward Dynamics Models},pdfauthor={Zipei Ma, Junzhe Jiang, Jiahui Zhang, Juntong Zhang, Chun Gu, Bozhou Zhang, Jiankang Deng, Xiatian Zhu, Li Zhang}}
\let\cite\citep

\newcommand{\model}{FDM}
\newcommand{\realtaskfigure}{\begin{figure}[!t]}
\newcommand{\fdmvlm}{\model{} \textnormal{\textit{w/}} VLM}
\newcommand{\fdmvgm}{\model{} \textnormal{\textit{w/}} VGM}

\title{Forward Dynamics Models}

\author{%
  \textbf{Zipei~Ma\textsuperscript{1,2}\quad Junzhe~Jiang\textsuperscript{1}\quad Jiahui~Zhang\textsuperscript{1}\quad Juntong~Zhang\textsuperscript{2}\quad Chun~Gu\textsuperscript{1,2}\quad Bozhou~Zhang\textsuperscript{1}\quad Jiankang~Deng\textsuperscript{3}\quad Xiatian~Zhu\textsuperscript{4}\quad Li~Zhang\textsuperscript{1,2}}\thanks{Corresponding author~\href{mailto:lizhangfd@fudan.edu.cn}{\texttt{lizhangfd@fudan.edu.cn}}.}\endgraf
  \mbox{\textsuperscript{1}School of Data Science, Fudan University}\quad
  \mbox{\textsuperscript{2}Shanghai Innovation Institute}\quad
  \mbox{\textsuperscript{3}Imperial College London}\quad
  \mbox{\textsuperscript{4}University of Surrey}\endgraf
  \vspace{.5em}
  \url{https://logosroboticsgroup.github.io/FDM/}
}

\makeatletter
\renewcommand{\@maketitle}{%
  \vbox{\hsize\textwidth
    {\LARGE\scshape\@title\par}
    \hspace*{\tabcolsep}%
    \parbox[t]{\dimexpr\textwidth-2\tabcolsep\relax}{%
      \raggedright\rule{0pt}{24pt}\@author}
    \vskip 0.3in minus 0.1in}}
\makeatother
\ClearShipoutPictureBG

\begin{document}
\maketitle
\suppressfloats[t]

\begin{abstract}

Vision-language-action (VLA) models and world action models (WAMs) have advanced robotic policy learning by building on pretrained vision-language models (VLMs) and video generative models (VGMs), respectively.
Yet VLAs learn without explicit supervision from future world states, while WAMs that condition actions on predicted futures incur extra inference cost and let visual prediction errors propagate into actions.
We present {\bf{\em Forward Dynamics Models}} (FDM), a family of robotic action models that couples action and future world-state prediction through an \emph{action-to-future} paradigm.
Future prediction is conditioned on action features, never the reverse: end-to-end training lets ground-truth future states supervise the policy through these features, while prediction errors cannot propagate into actions.
At inference, FDM skips future generation entirely, avoiding its computational cost.
The paradigm is backbone-agnostic, and we instantiate it with both VLM and VGM backbones.
Experiments in simulation and on real robots demonstrate the effectiveness of FDM across both backbone families, achieving a state-of-the-art \textbf{99.4\%} average success rate on LIBERO and absolute gains of up to \textbf{13\%} over VLA and WAM methods on real-world tasks.

\end{abstract}

\section{Introduction}

\begin{figure*}[t]
    \centering
    \includegraphics[width=\linewidth]{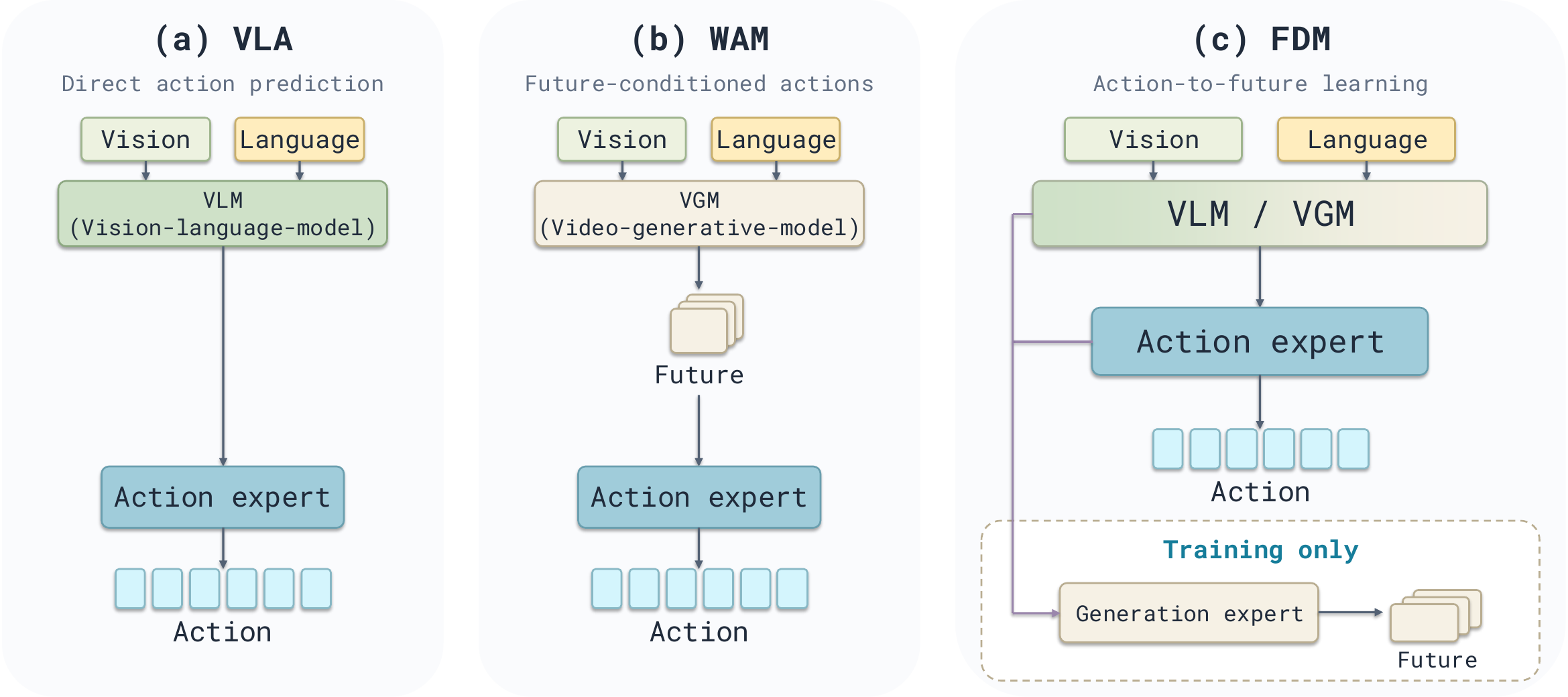}
    \caption{\textbf{Architectural comparison of action prediction and future supervision.}
    \textbf{(a)} VLA predicts actions directly from the current observation and instruction.
    \textbf{(b)} A future-conditioned WAM conditions action prediction on predicted future video, requiring future generation at inference.
    \textbf{(c)} \model{} conditions future prediction on context and action features through \textit{action-to-future} learning.
    During training, action imitation and future prediction are jointly optimized; the future loss backpropagates through action features to supervise the policy.
    At inference, \model{} denoises action chunks using only the current observation and instruction, skipping the future-generation branch marked as training only while retaining context processing.}
    \label{fig:pipe_comp}
\end{figure*}

Frontier vision-language models (VLMs) and video generative models (VGMs) offer two complementary foundations for robotic action learning.
Vision-language-action (VLA) models adapt VLMs to map current observations and task instructions directly to actions~\cite{kim24openvla, intelligence2025pi_, yu2026wall, GR00T}, following the action-imitation paradigm~\cite{Diffusionpolicy, zhao2023act} (\hyperref[fig:pipe_comp]{\Cref*{fig:pipe_comp}(a)}).
While effective, this paradigm learns \emph{what} to do without explicit supervision on \emph{what happens next}, leaving the rich signal in future world states unused.
World action models (WAMs) instead build on video backbones to model visual dynamics alongside actions~\cite{hu2024video, feng2025vidar, pai2025mimic, bi2026motus}.
A prevalent design first predicts future world states and then infers actions from them in an inverse-dynamics manner~\cite{li2026wall, zhao2025cot, ye2026worldactionmodelszeroshot, ma2026internvla, agibot2026geact2, tian2024seer} (\hyperref[fig:pipe_comp]{\Cref*{fig:pipe_comp}(b)}).
However, future prediction is inherently error-prone, since every visual detail matters across both space and time.
Placing it \emph{before} action prediction means its errors accumulate and amplify into the actions, and future generation also adds substantial inference overhead.
Other works predict future frames and actions jointly or in parallel~\cite{ye2026worldactionmodelszeroshot, wang2025unified}, align policy features with future embeddings~\cite{zheng2025flare}, or train an action model and an action-conditioned world model side by side~\cite{cen2025worldvla}.
These designs motivate studying the specific contribution of future supervision through action features, beyond supervision through shared context.

We investigate whether a policy can benefit from future supervision without ever depending on predicted futures.
To this end, we introduce \textbf{\em Forward Dynamics Models} (\model{}), a family of robotic action models built on an \emph{action-to-future} learning paradigm.
Borrowing the classical notion of forward dynamics~\cite{jordan1992forward, agrawal2016poke}, the mapping from state and action to the next state, \model{} connects action learning with future world state prediction:
given the current world state and task instruction, \model{} jointly predicts actions and future world states with asymmetric conditioning.
Future prediction reads action features, while action prediction cannot access future information.
The model is trained end to end, so the future-prediction loss backpropagates through these action features into the policy.
Action-to-future specifies this one-way information dependency within joint prediction.

\begin{wrapfigure}{r}{0.50\textwidth}
    \vspace{-\intextsep}
    \centering
    \begingroup
    \setlength{\unitlength}{1bp}
    \resizebox{\linewidth}{!}{%
        \begin{picture}(518.4,403.2)
            \put(0,0){\includegraphics[width=518.4bp]{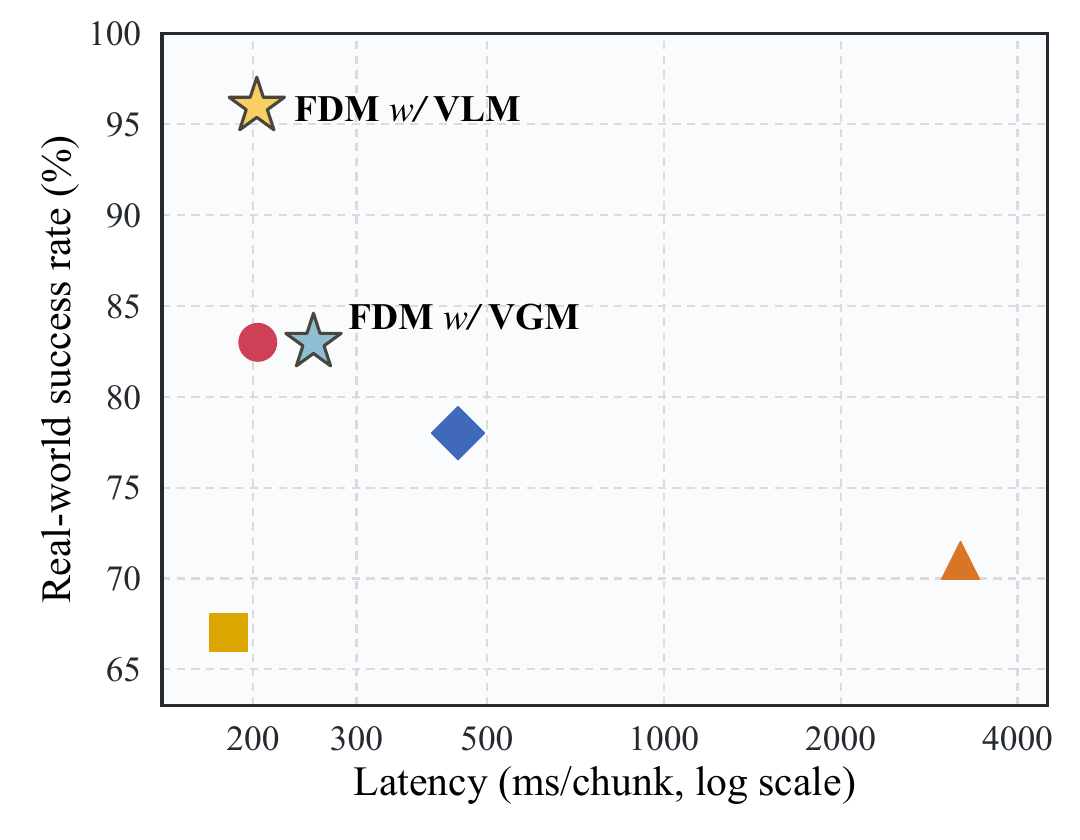}}
            \fontsize{18}{22}\selectfont
            \put(94.74,218.87){\makebox(0,0)[l]{\scalebox{2.4}{\normalsize $\pi_{0.5}$}~\citep{intelligence2025pi_}}}
            \put(219.85,168.28){\makebox(0,0){FastWAM-Joint~\citep{yuan2026fastwam}}}
            \put(134.86,99.38){\makebox(0,0)[l]{\scalebox{2.4}{\normalsize $\pi_0$}~\citep{black2024pi_0}}}
            \put(444.05,134.25){\makebox(0,0)[r]{DreamZero~\citep{ye2026worldactionmodelszeroshot}}}
        \end{picture}%
    }
    \endgroup
    \caption{\textbf{Real-world success rate vs. latency.}
    Success rates are averaged over five manipulation tasks on a dual-arm ARX-X5 robot: drawer packing, box packing and closing, shuttlecock insertion, paper-cup stacking, and block stacking, with 20 trials per method per task.
    Inference latency is measured on one NVIDIA H100 GPU.}
    \label{fig:success_latency}
\end{wrapfigure}

This reverses the conditioning direction of future-conditioned WAMs and brings several advantages:
(1) \emph{Robustness}: action prediction does not depend on predicted future world states, so errors in those predictions cannot propagate into actions;
(2) \emph{Action-grounded supervision}: ground-truth future states supervise the action features used to generate actions, complementing action imitation with supervision from observed consequences;
(3) \emph{Efficiency}: action-only inference skips future generation while retaining the context processing required for action prediction;
(4) \emph{Flexibility}: either a VLM or a VGM can serve as the backbone.
As shown in \Cref{fig:success_latency}, \fdmvlm{} improves real-world success at comparable latency to $\pi_{0.5}$, while \fdmvgm{} improves success with $1.8\times$ faster inference than FastWAM-Joint.
\Cref{tab:para_compare} contrasts the formulations of the three paradigms.
Our analysis further confirms that, unlike FastWAM-Joint, whose action error grows with the quality degradation of predicted futures, \model{}'s action error is insensitive to future-prediction quality (\Cref{sec:err_prop}).

\begin{table}[tb]
    \caption{\textbf{Formulation comparison.}
    $C=(O_t,L)$: input context comprising the current observation $O_t$ and task instruction $L$; $A$: action chunk; $V$: future world state target.
    WAM here denotes the future-conditioned formulation.
    For FDM, the formulation expresses action-to-future dependencies realized through action features; inference requires only action prediction.
    }
    \label{tab:para_compare}
    \begin{center}
    \small
    \setlength{\tabcolsep}{8pt}
    \renewcommand{\arraystretch}{1.3}
    \begin{tabular}{c|ll}
        \toprule
        \textbf{Model family} & \textbf{Training} & \textbf{Inference} \\
        \midrule
        VLA & $p_\theta(A \mid C)$ & $p_\theta(A \mid C)$ \\
        WAM & $p_\theta(V \mid C)\,p_\theta(A \mid C,V)$ & $p_\theta(V \mid C)\,p_\theta(A \mid C,V)$ \\
        \model{} & $p_\theta(A \mid C)\,p_\theta(V \mid C,A)$ & $p_\theta(A \mid C)$ \\
        \bottomrule
    \end{tabular}
    \end{center}
\end{table}

Our \textit{contributions} are threefold:
\textbf{(i)} We introduce \model{}, a novel family of robotic action models that learns in an action-to-future manner during training, with future supervision flowing into the policy through its action features, and requires no future prediction at inference;
\textbf{(ii)} We build VLM- and VGM-based instantiations of \model{}, demonstrating its backbone flexibility;
\textbf{(iii)} Experiments in simulation and on real robots show that \model{} achieves strong or competitive performance across VLM- and VGM-based settings, and analysis shows that it avoids the error propagation of future-conditioned designs.

\section{Related work}

\textbf{Vision-language-action models.}
VLA models adapt pretrained vision-language models to predict robot actions~\cite{wang2026qwen, yuan2026qwen, ye2026starvla, black2024pi_0, yang2026abot, liu2026streampi, yu2026affordancevla, zhang2026onevla, zheng2026x, qu2025spatialvla, internvlam1, internvla_a1, shukor2025smolvla, wang2025vlaadapter}.
Recent approaches also predict future observations or their latent representations~\cite{cen2025worldvla, ma2026internvla, cheang2024gr, cen2025rynnvla, lv2025f1, zhang2026uam, hu2026bagelvla, kang2026xtokenizer, yang2025mantis, zhang2025upvla}, using future prediction as a learning objective or an input to action generation, as in visual chain-of-thought policies~\cite{zhao2025cot}.
Our \fdmvlm{} builds on $\pi_{0.5}$ and conditions future prediction on action features and the current context, establishing an action-to-future dependency.

\textbf{World action models.}
Video models provide visual dynamics representations for robot control~\cite{hacohen2024ltx, ali2025world, wan2025wan}.
WAMs use these representations to condition action decoding~\cite{hu2024video, pai2025mimic, ma2026dit4dit} or jointly predict video and actions~\cite{bi2026motus, ye2026worldactionmodelszeroshot, lingbot-va2026, wang2026openwam, zhou2026zerowam, tian2026vtwam, zhu2025uwm}.
Extensions combine predicted outcomes with value estimates for planning~\cite{bi2026motus2, kim2026cosmos, chen2026mvwam}, while asynchronous designs improve efficiency through different denoising budgets or prediction horizons~\cite{guo2026xwam, cai2026aha}.
Our \fdmvgm{} places future prediction downstream of action features, keeping the action path independent of predicted video.
Deployment therefore omits future-video computation while retaining current-observation processing.

Joint predictive models use visual supervision for action learning~\cite{wang2025unified, ma2026geosem, ye2026gigaworld, xu2026nextforcing, wu2023gr1, zheng2025flare, won2025dust}.
UVA and Fast-WAM also enable action-only deployment via separate video--action decoders and structured attention, respectively~\cite{li2503unified, yuan2026fastwam}.
FDM conditions future prediction on action features and backpropagates prediction errors into them.
Our context-only and stop-gradient controls isolate action conditioning and backward supervision, respectively.

\section{Method}
\label{sec:method}

\begin{figure*}[t]
    \centering
    \includegraphics[width=\linewidth]{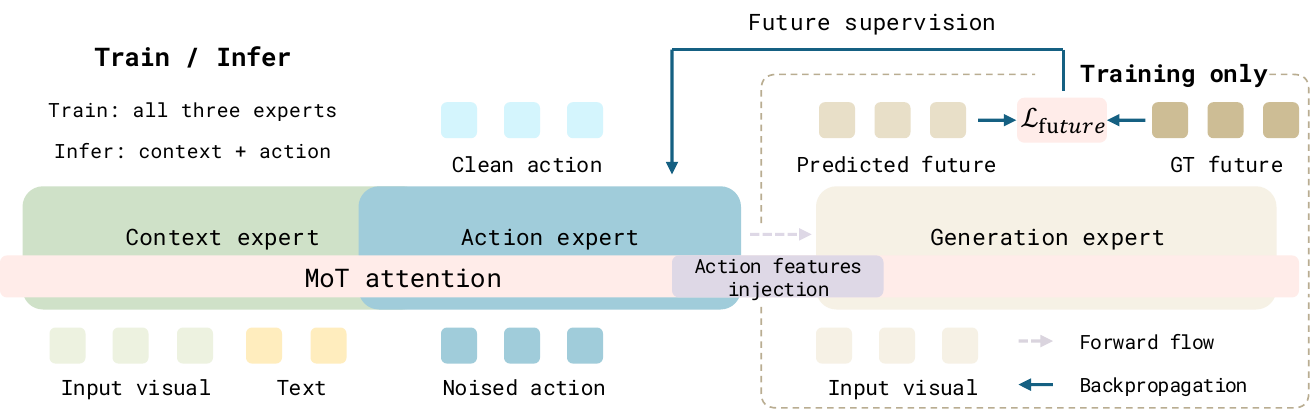}
    \caption{\textbf{Overview of our Forward Dynamics Models.}
    The context expert encodes visual observations and the instruction to guide the action expert.
    During training, the generation expert predicts future states from context and action features; its loss backpropagates through these action features to complement action imitation.
    This asymmetric connection lets future supervision shape the policy while keeping action prediction independent of predicted futures.
    At inference, the action expert generates actions using the required context features, skipping future generation.}
    \label{fig:pipeline}
\end{figure*}

FDM is a family of robotic action models that jointly predicts actions and future world states.
Its action-to-future paradigm uses asymmetric conditioning: future prediction reads action features, while action prediction cannot access future information.
Future prediction errors supervise action learning through these features during training.
Let $C=(O_t,L)$ denote the current observation $O_t$ and task instruction $L$, $A$ an action chunk, and $V$ the future world state target.
The current world state is observed through images; future world state targets are represented by VAE latents of future images or video in our implementations.
The intended information dependencies are
\begin{equation}
\label{eq:model_factorization}
p\left(A, V \mid C\right)
= p\left(A \mid C\right) p\left(V \mid C, A\right).
\end{equation}
\Cref{eq:model_factorization} is a conceptual statement of these dependencies; the learning mechanism is the representation-level connection defined below.
In our implementations, the future-prediction branch is conditioned through action features (the action expert's hidden states), rather than through the final numerical action output.
This conditioning direction holds within joint prediction and permits action-only inference that skips future generation.
Future prediction remains optional for visual rollout.
We first specify the action features and supervision path, then describe how our VLM- and VGM-based instantiations realize it.

\subsection{Action-to-future features and supervision}
\label{sec:supervision}

Let $H_a$ denote the action features read by the future-prediction branch, and let $\theta_a$ denote the parameters that produce them.
In our implementations, $H_a$ consists of action-expert hidden states, and $\theta_a$ denotes the action-expert parameters.
Both instantiations use action flow matching, with $X_a=(A_{\tau_a},\tau_a)$ comprising the noised action chunk $A_{\tau_a}$ and its action flow-matching time $\tau_a\in[0,1]$.
The action velocity prediction and future objective take the form
\begin{equation}
\label{eq:representation_interface}
H_a=f_{\theta_a}(C,X_a),\qquad
\widehat u_a=D_\psi(H_a),\qquad
\mathcal{L}_{\mathrm{world}}=\ell(G_\phi(C,H_a,\ldots),V),
\end{equation}
where $D_\psi$ is the action velocity head and $\ell$ denotes the appropriate future-prediction objective, with noise and time dependencies suppressed in this schematic expression.
The generation expert reads $H_a$; the velocity output $\widehat u_a$ is used to update the action chunk during denoising.
For \fdmvlm{}, $G_\phi$ predicts future-image latents using the MSE objective in \Cref{eq:latent_mse}; for \fdmvgm{}, it is the velocity predictor $v_\phi$ trained using the flow-matching objective in \Cref{eq:video_loss}.
The future objective $\mathcal{L}_{\mathrm{world}}$ contributes the gradient
\begin{equation}
\label{eq:future_action_gradient}
\left.\nabla_{\theta_a}\mathcal{L}_{\mathrm{world}}\right|_{H_a}
=\left(\frac{\partial H_a}{\partial\theta_a}\right)^{\!\top}
\nabla_{H_a}\mathcal{L}_{\mathrm{world}}.
\end{equation}
This path supervises the internal features used for action generation.
To isolate its contribution, \Cref{sec:ablation} replaces only the generation-facing $H_a$ with $\operatorname{sg}(H_a)$, which preserves its values but blocks gradients, and measures the change in closed-loop policy success.
Future-latent perturbations examine error propagation to actions at inference, while gradient checks examine the backward supervision signal.

Our two instantiations implement these dependencies with a mixture-of-transformers (MoT) architecture~\cite{liang2024mixture} (\Cref{fig:pipeline}).
Each expert processes its tokens with self-attention, while MoT connects the experts through their hidden states.
The generation expert reads action features as conditions for future prediction; the action expert cannot read future visual states.
This restriction also applies to indirect paths: context features accessible to the action expert must remain independent of generation-branch tokens (\Cref{fig:supp_attention_mask}).
Thus, future supervision can train the policy without making future prediction an input to action generation.

\subsection{\fdmvlm{}}
\label{sec:instantiations}

\fdmvlm{}, our VLM-based instantiation, uses a context expert, an action expert, and a generation expert.
The context expert plays the role of the VLM in a conventional VLA, extracting semantic features from the current images and instruction.
The action expert reads these context features to generate actions.
The generation expert combines the Wan VAE latents~\cite{wan2025wan} of historical and current images with context features and action features to directly predict the clean latent of one future image per camera view.

Let $V$ denote the future-image latents across camera views and $\widehat V=G_\phi(C,H_a,\ldots)$ their prediction.
We minimize the mean squared error between predicted and target latents:
\begin{equation}
\label{eq:latent_mse}
\mathcal{L}_{\mathrm{img}}
=\mathbb{E}\left[\operatorname{MSE}(\widehat V,V)\right],
\end{equation}
where $\operatorname{MSE}$ averages the per-view mean squared errors across camera views.

\subsection{\fdmvgm{}}
\label{sec:wam}

\fdmvgm{}, our VGM-based instantiation, uses an action expert and a generation expert.
The video backbone serves both the context-processing and generation roles: it extracts current-observation features for action generation and predicts action-conditioned future video.
The action expert reads the generation expert's current-observation features, while future-video tokens read action features through MoT.
Although observation and future tokens belong to the same generation expert, their attention permissions differ: the features available to the action policy cannot incorporate future information.

Here, $V$ denotes the clean future-video latents encoded by the Wan VAE, and $C$ provides the current observation and task instruction.
During training, the current-frame latent remains fixed while the future latents are noised for flow matching.
Using the convention
\begin{equation}
\label{eq:fm}
V_{\tau_v}=(1-\tau_v)V+\tau_v\epsilon_v,
\qquad \epsilon_v\sim\mathcal{N}(0,I),
\end{equation}
the generation expert predicts a velocity field and minimizes
\begin{equation}
\label{eq:video_loss}
\mathcal{L}_{\mathrm{vid}}
=\mathbb{E}\left[\left\|
 v_\phi(V_{\tau_v},\tau_v;C,H_a)-(\epsilon_v-V)
\right\|_2^2\right],
\end{equation}
where $\tau_v\in[0,1]$ is the video flow-matching time, independent of the action flow-matching time $\tau_a$, and the loss is evaluated only on future latents.
Future-video tokens interact through video attention to model temporal content.

\subsection{Model training and inference}
\label{sec:training}

Both instantiations combine the action flow-matching objective $\mathcal{L}_{\mathrm{act}}$ with future supervision:
\begin{equation}
\label{eq:total}
\mathcal{L}_{\mathrm{total}}
=\lambda_a\mathcal{L}_{\mathrm{act}}
+\lambda_v\mathcal{L}_{\mathrm{world}},
\end{equation}
where $\lambda_a$ and $\lambda_v$ weight action learning and future supervision, respectively, and $\mathcal{L}_{\mathrm{world}}$ is $\mathcal{L}_{\mathrm{img}}$ for \fdmvlm{} and $\mathcal{L}_{\mathrm{vid}}$ for \fdmvgm{}.
Real-world training settings are provided in \Cref{sec:supp_hyperparameters}.
The action objective trains the policy to reproduce demonstrated actions, while the future objective supervises the action-expert features through their MoT connection to the generation expert.

\paragraph{Action-only inference.}
\label{sec:inference}
At inference, both policies generate action chunks from noise conditioned on the current observation and task instruction, without predicting future visual states.
This action-only path avoids compounding errors from future visual predictions and removes the computational overhead of future generation, supporting high-frequency reactions to new observations.

\begin{table*}[!t]
    \setlength{\belowcaptionskip}{6pt}
    \begin{minipage}[t]{0.54\textwidth}
    \centering
    \caption{\textbf{Results on LIBERO.} Success rates (\%). P.T.: robot policy pretraining. Ranks within each backbone family (\Crefrange{tab:libero}{tab:robotwin}): \colorbox{best_result}{best} / \colorbox{second_result}{second best}.}
    \label{tab:libero}
    \footnotesize
    \renewcommand{\arraystretch}{1.12}
    \setlength{\tabcolsep}{3pt}
    \resizebox{\linewidth}{!}{%
    \begin{tabular}{@{}lc|cccc|c@{}}
    \toprule
    \textbf{Method} & P.T. & Spatial & Object & Goal & Long & \textbf{Avg.} \\
    \midrule
    \multicolumn{7}{@{}l}{\textit{VLM-based}} \\
    $\pi_0$~\cite{black2024pi_0} & \cmark & 96.8 & \cellcolor{second_result}98.8 & 95.8 & 85.2 & 94.1 \\
    $\pi_{0.5}$~\cite{intelligence2025pi_} & \cmark & \cellcolor{second_result}98.8 & 98.2 & \cellcolor{second_result}98.0 & \cellcolor{second_result}92.4 & \cellcolor{second_result}96.9 \\
    \fdmvlm{} (Ours) & \xmark & \cellcolor{best_result}99.2 & \cellcolor{best_result}100.0 & \cellcolor{best_result}99.8 & \cellcolor{best_result}98.6 & \cellcolor{best_result}99.4 \\
    \midrule
    \multicolumn{7}{@{}l}{\textit{VGM-based}} \\
    Motus~\cite{bi2026motus} & \cmark & 96.8 & \cellcolor{second_result}99.8 & 96.6 & \cellcolor{second_result}97.6 & 97.7 \\
    LingBot-VA~\cite{lingbot-va2026} & \cmark & \cellcolor{second_result}98.5 & 99.6 & \cellcolor{second_result}97.2 & \cellcolor{best_result}98.5 & \cellcolor{best_result}98.5 \\
    \fdmvgm{} (Ours) & \xmark & \cellcolor{best_result}98.8 & \cellcolor{best_result}100.0 & \cellcolor{best_result}99.2 & 94.2 & \cellcolor{second_result}98.1 \\
    \bottomrule
    \end{tabular}%
    }
    \end{minipage}\hfill
    \begin{minipage}[t]{0.44\textwidth}
    \ifdefined\capstart\capstart\fi
    \centering
    \caption{\textbf{Results on RoboTwin 2.0.} Success rates (\%) over 50 tasks. Avg. averages Clean and Rand.\vphantom{\colorbox{best_result}{best}}}
    \label{tab:robotwin}
    \footnotesize
    \renewcommand{\arraystretch}{1.12}
    \setlength{\tabcolsep}{3.428pt}
    \resizebox{\linewidth}{!}{%
    \begin{tabular}{@{}lc|cc|c@{}}
    \toprule
    \textbf{Method} & P.T. & Clean & Rand. & \textbf{Avg.} \\
    \midrule
    \multicolumn{5}{@{}l}{\textit{VLM-based}} \\
    $\pi_0$~\cite{black2024pi_0} & \cmark & 65.92 & 58.40 & 62.16 \\
    $\pi_{0.5}$~\cite{intelligence2025pi_} & \cmark & \cellcolor{best_result}82.74 & \cellcolor{second_result}76.76 & \cellcolor{second_result}79.75 \\
    \fdmvlm{} (Ours) & \xmark & \cellcolor{second_result}81.54 & \cellcolor{best_result}80.86 & \cellcolor{best_result}81.20 \\
    \midrule
    \multicolumn{5}{@{}l}{\textit{VGM-based}} \\
    Motus~\cite{bi2026motus} & \cmark & 88.66 & 87.02 & 87.8 \\
    LingBot-VA~\cite{lingbot-va2026} & \cmark & \cellcolor{best_result}92.90 & \cellcolor{best_result}91.50 & \cellcolor{best_result}92.20 \\
    \fdmvgm{} (Ours) & \xmark & \cellcolor{second_result}88.84 & \cellcolor{second_result}89.62 & \cellcolor{second_result}89.23 \\
    \bottomrule
    \end{tabular}%
    }

    \end{minipage}
    \vspace{-10pt}
\end{table*}

\section{Experiments}
\label{sec:experiments}

We evaluate \model{} across two backbone families in simulation and the real world.
Beyond task success, we analyze how future supervision contributes to action learning and how the action-to-future design affects inference behavior and computational cost.

\subsection{Experimental setup}
\label{sec:setup}

Training uses 8 NVIDIA H100 GPUs, and all inference runs on 1 NVIDIA H100 GPU.
Appendix~\ref{sec:supp_hyperparameters} reports the training hyperparameters for \fdmvlm{} and \fdmvgm{} used in the real-world experiments.

\paragraph{Two backbone families.}
We instantiate \model{} with two backbone families to assess future supervision across different policy architectures.
For \fdmvlm{}, we extend $\pi_{0.5}$~\cite{intelligence2025pi_} with a randomly initialized generation expert that has the same network architecture as its action expert.
For \fdmvgm{}, we add a randomly initialized action expert to Wan2.2-5B~\cite{wan2025wan}.
In both instantiations, MoT~\cite{liang2024mixture} enables the future-prediction branch to read action features, following the directional information flow in \Cref{sec:method}.

\providecommand{\realtaskfigure}{\begin{figure}[H]}
\realtaskfigure
    \centering
    \begin{minipage}[t]{0.192\linewidth}
        \centering
        {\footnotesize\strut Drawer packing}\par\vspace{3pt}
        \includegraphics[width=\linewidth]{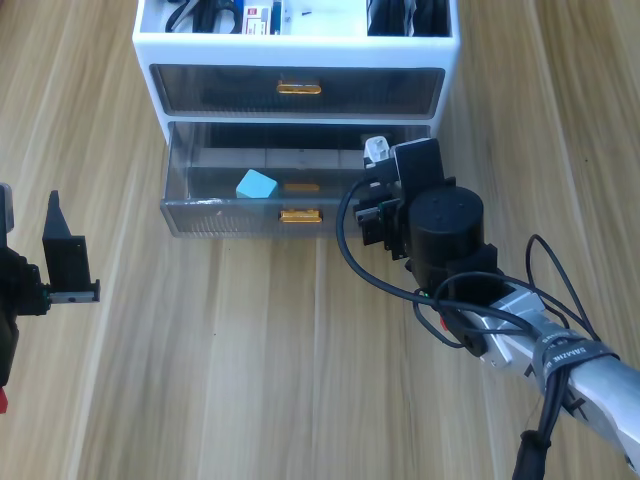}\par\vspace{3pt}
        \includegraphics[width=\linewidth]{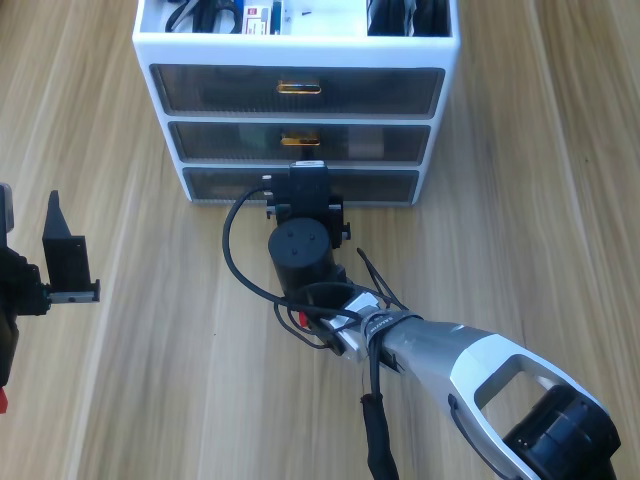}
    \end{minipage}\hfill%
    \begin{minipage}[t]{0.192\linewidth}
        \centering
        {\footnotesize\strut Box packing}\par\vspace{3pt}
        \includegraphics[width=\linewidth]{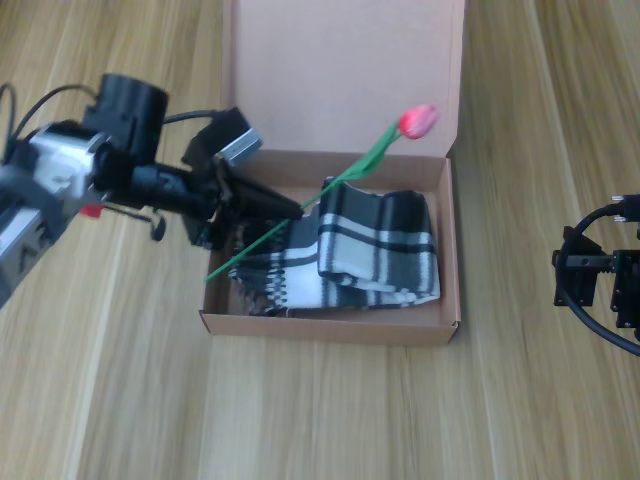}\par\vspace{3pt}
        \includegraphics[width=\linewidth]{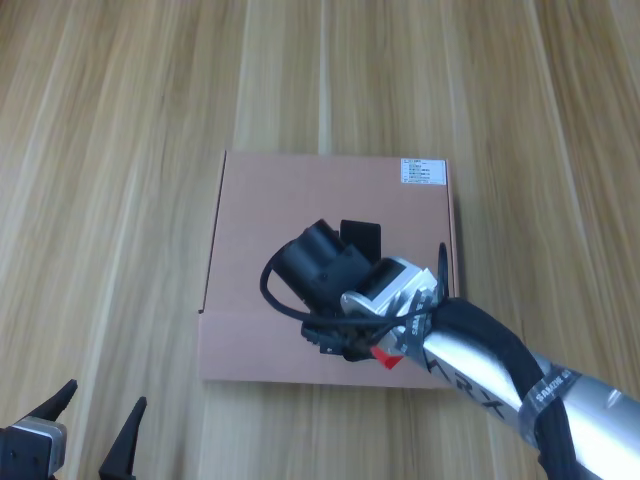}
    \end{minipage}\hfill%
    \begin{minipage}[t]{0.192\linewidth}
        \centering
        {\footnotesize\strut Shuttlecock insertion}\par\vspace{3pt}
        \includegraphics[width=\linewidth]{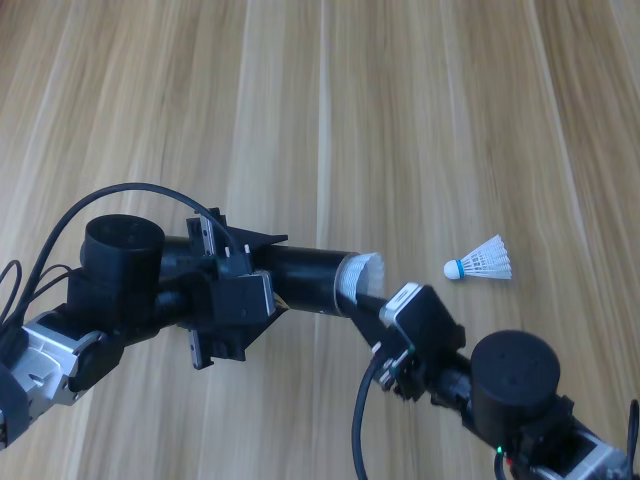}\par\vspace{3pt}
        \includegraphics[width=\linewidth]{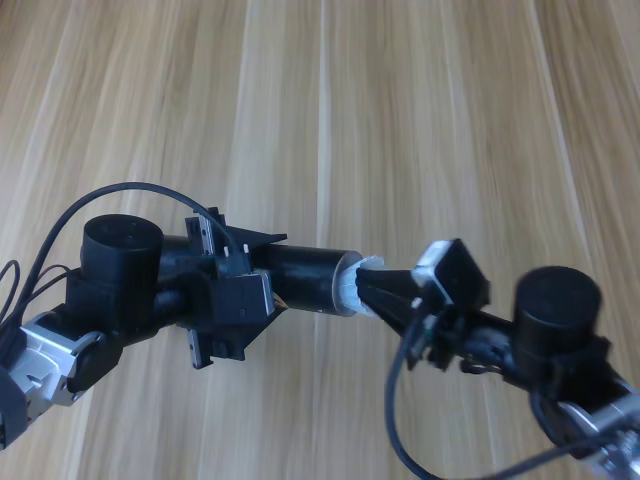}
    \end{minipage}\hfill%
    \begin{minipage}[t]{0.192\linewidth}
        \centering
        {\footnotesize\strut Cup stacking}\par\vspace{3pt}
        \includegraphics[width=\linewidth]{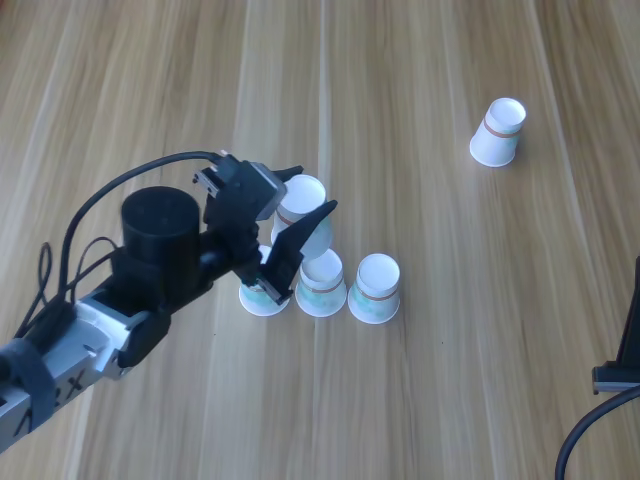}\par\vspace{3pt}
        \includegraphics[width=\linewidth]{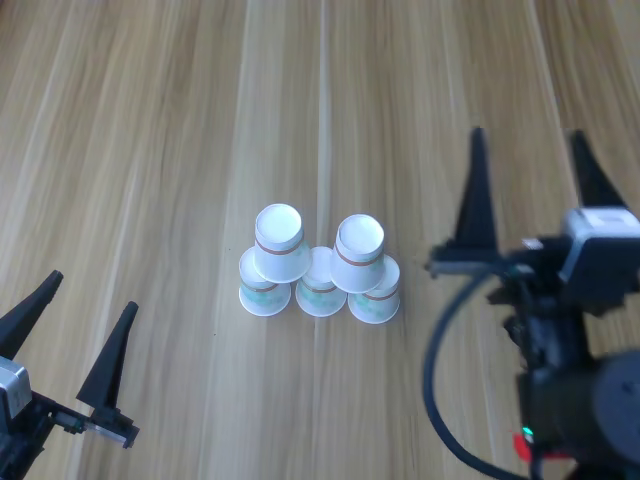}
    \end{minipage}\hfill%
    \begin{minipage}[t]{0.192\linewidth}
        \centering
        {\footnotesize\strut Block stacking}\par\vspace{3pt}
        \includegraphics[width=\linewidth]{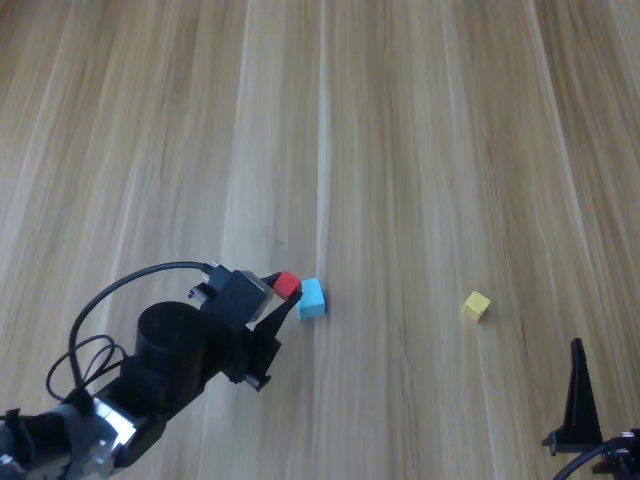}\par\vspace{3pt}
        \includegraphics[width=\linewidth]{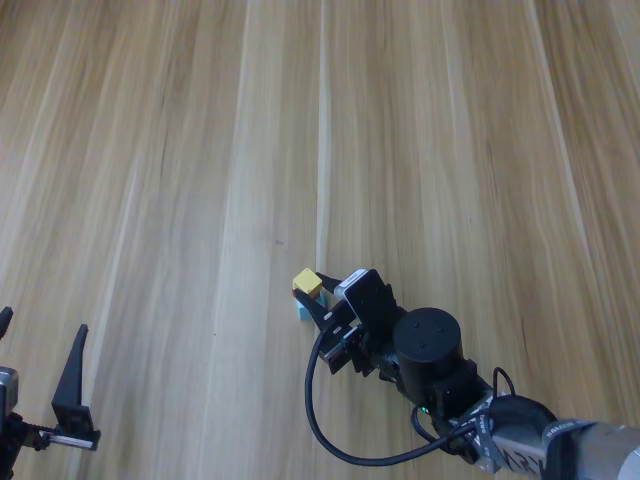}
    \end{minipage}
    \caption{\textbf{Real-world task examples.} Each column shows two frames from one task, ordered chronologically from top to bottom: drawer packing, box packing and closing, shuttlecock insertion, cup stacking, and block stacking.}
    \label{fig:real_task_examples}
\end{figure}

\paragraph{Benchmarks and real-world tasks.}
We evaluate task success in simulation on LIBERO~\cite{liu2023libero} and RoboTwin 2.0~\cite{chen2025robotwin}.
LIBERO covers the Spatial, Object, Goal, and Long suites, while RoboTwin evaluates bimanual manipulation under clean and randomized conditions.
On both simulation benchmarks, we compare \fdmvlm{} with the VLA methods $\pi_0$ and $\pi_{0.5}$, and \fdmvgm{} with the WAM methods Motus~\cite{bi2026motus} and LingBot-VA~\cite{lingbot-va2026}.
Both \model{} instantiations are trained without robot policy pretraining (P.T.).
To assess performance beyond simulation, we use a dual-arm ARX-X5 robot for five tasks: loading a drawer with a charger and a block, then closing it (Drawer); packing a scarf and a flower into a box and closing it (Box); inserting two shuttlecocks into a tube (Shuttlecock); stacking two cups on a three-cup base (Cup); and stacking two small blocks on a larger block (Block).
\Cref{fig:real_task_examples} illustrates the five real-world tasks with two frames per task.
The methods compared in the real world are $\pi_0$ and $\pi_{0.5}$ for the VLM-based model, and DreamZero and FastWAM-Joint (a future-conditioned WAM) for the VGM-based model.
FastWAM-Joint shares the architecture of \fdmvgm{}, differing only in its attention mask, which reverses the direction of cross-expert attention.
We use the same FastWAM-Joint implementation for real-world task success (\Cref{sec:sim_results}), future-perturbation analysis (\Cref{sec:future_error_sensitivity}), and efficiency evaluation (\Cref{sec:efficiency}).
The hardware setup, teleoperation interface, and detailed task descriptions are provided in Appendix~\ref{sec:supp_real_tasks}.

\subsection{Q1: Does future supervision improve task success?}
\label{sec:sim_results}

We assess the effectiveness of future supervision by comparing each \model{} instantiation with VLA and WAM methods, first on LIBERO and RoboTwin and then on five real-robot tasks.
This evaluation examines whether the benefit holds across policy architectures and manipulation settings.

\paragraph{LIBERO.}
\fdmvlm{} improves over $\pi_{0.5}$ across all four suites, raising average success from 96.9\% to 99.4\% and LIBERO-Long success from 92.4\% to 98.6\% (\Cref{tab:libero}).
\fdmvgm{} achieves 98.1\% average success, compared with 97.7\% for Motus and 98.5\% for LingBot-VA.

\paragraph{RoboTwin.}
\fdmvlm{} and \fdmvgm{} achieve average success rates of 81.20\% and 89.23\%, respectively, improving over $\pi_{0.5}$ (79.75\%) and Motus (87.8\%) (\Cref{tab:robotwin}).
Both improve under randomized conditions.
Appendix~\ref{sec:supp_robotwin_results} reports results for all 50 tasks.

\paragraph{Real-world manipulation.}
\label{sec:real_robot}

We next evaluate whether the gains observed in simulation carry over to physical manipulation.
\fdmvlm{} achieves 96\% average success, improving over $\pi_{0.5}$ (83\%) on all five tasks, while \fdmvgm{} reaches 83\%, compared with 78\% for the architecture-matched FastWAM-Joint and 71\% for DreamZero (\Cref{fig:real_future}).
Both instantiations achieve 100\% success on Cup and Block, demonstrating strong performance on the two stacking tasks within an evaluation that also covers sequential object handling, cloth manipulation, and insertion.

\begin{figure*}[!t]
    \centering
    \includegraphics[width=\linewidth]{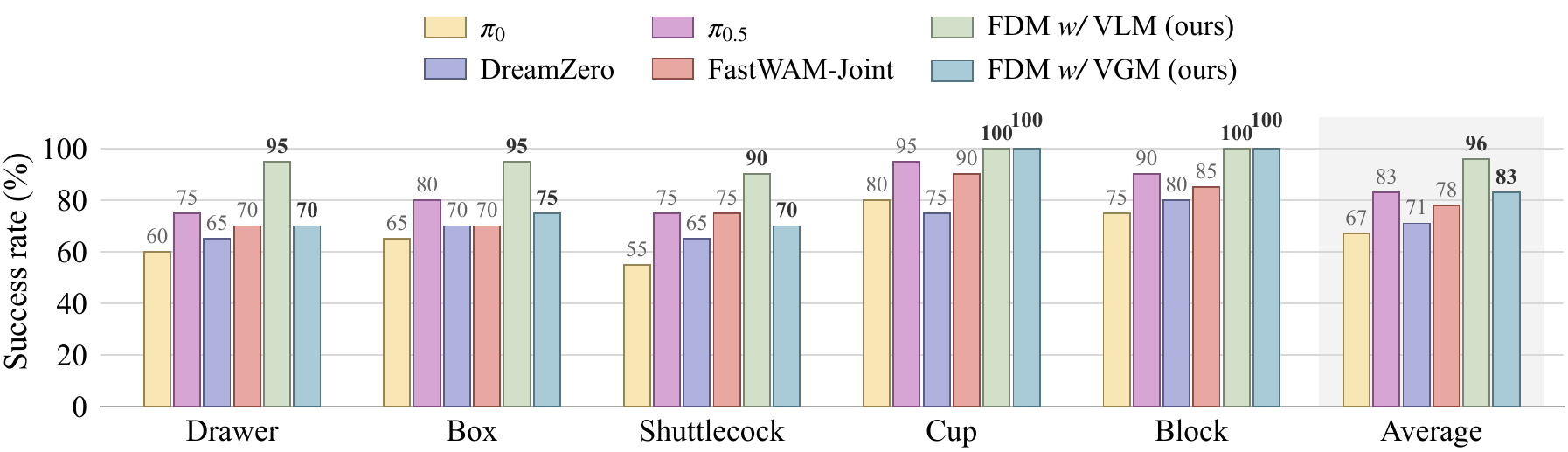}
    \caption{\textbf{Main results on the five real-robot tasks.}
    Per-task and average success rates, 20 trials per method per task.
    Both FDM variants outperform the corresponding comparison methods on average.}
    \label{fig:real_future}
    \vspace{-10pt}
\end{figure*}

\subsection{Q2: Does backpropagation through action features improve control?}
\label{sec:ablation}

Backpropagating future prediction errors through $H_a$ improves closed-loop task success in both backbone families.
To isolate this contribution, we ablate the action-to-future connection on five real-robot tasks.
We compare four variants that separate the presence of a future objective, forward access to action features, and backward supervision through that interface (\Cref{tab:ablation}).
\textbf{A. Action-only} trains with action imitation alone.
\textbf{B. Context-only future} adds the future objective but prevents generation-branch tokens from reading action features.
\textbf{C. Stop-gradient action conditioning} lets the future-prediction branch read $\operatorname{sg}(H_a)$, which preserves values but blocks gradients.
\textbf{D. Full \model{}} lets the future-prediction branch read $H_a$ and backpropagate through it.

\paragraph{Controlled protocol.}
Within each backbone, we compare all four variants under matched training and evaluation settings, using the same training seeds.
We measure closed-loop success on the five real-robot tasks with action-only inference.
Appendix~\ref{sec:supp_ablation_protocol} details the controlled settings and stop-gradient implementation.

\begin{table*}[t]
    \caption{\textbf{Action-to-future ablation on five real-robot tasks.}
    Future: future supervision; Reads $H_a$: action-feature conditioning; Backprop: future-loss gradients through $H_a$.}
    \label{tab:ablation}
    \begin{center}
    \small
    \setlength{\tabcolsep}{1.8pt}
    \renewcommand{\arraystretch}{1.15}
    \begin{tabular}{@{}llccccccccc@{}}
        \toprule
        & & \multicolumn{3}{c}{Action-to-future pathway} & \multicolumn{6}{c}{Success rate (\%) $\uparrow$} \\
        \cmidrule(lr){3-5} \cmidrule(lr){6-11}
        Model & Variant & Future & Reads $H_a$ & Backprop & Drawer & Box & Shuttlecock & Cup & Block & Average \\
        \midrule
        \multirow{4}{*}{VLM} & A. Action-only & \xmark & \xmark & \xmark & 75 & 80 & 75 & \cellcolor{second_result}95 & 90 & 83 \\
         & B. Context-only future & \cmark & \xmark & \xmark & \cellcolor{second_result}85 & 80 & 80 & \cellcolor{second_result}95 & \cellcolor{second_result}95 & 87 \\
         & C. Stop-gradient & \cmark & \cmark & \xmark & 80 & \cellcolor{second_result}85 & \cellcolor{second_result}85 & \cellcolor{second_result}95 & \cellcolor{best_result}100 & \cellcolor{second_result}89 \\
         & D. \model{} (Ours) & \cmark & \cmark & \cmark & \cellcolor{best_result}95 & \cellcolor{best_result}95 & \cellcolor{best_result}90 & \cellcolor{best_result}100 & \cellcolor{best_result}100 & \cellcolor{best_result}96 \\
        \midrule
        \multirow{4}{*}{VGM} & A. Action-only & \xmark & \xmark & \xmark & 50 & 60 & 55 & 85 & 80 & 66 \\
         & B. Context-only future & \cmark & \xmark & \xmark & \cellcolor{second_result}60 & 65 & 55 & \cellcolor{second_result}90 & 90 & 72 \\
         & C. Stop-gradient & \cmark & \cmark & \xmark & \cellcolor{best_result}70 & \cellcolor{second_result}70 & \cellcolor{second_result}65 & \cellcolor{second_result}90 & \cellcolor{second_result}95 & \cellcolor{second_result}78 \\
         & D. \model{} (Ours) & \cmark & \cmark & \cmark & \cellcolor{best_result}70 & \cellcolor{best_result}75 & \cellcolor{best_result}70 & \cellcolor{best_result}100 & \cellcolor{best_result}100 & \cellcolor{best_result}83 \\
        \bottomrule
    \end{tabular}
    \vspace{-10pt}
    \end{center}
\end{table*}

\paragraph{Backward future supervision improves control (D vs. C).}
Full \model{} raises average success from 89\% to 96\% for the VLM-based model and from 78\% to 83\% for the VGM-based model, gains of 7 and 5 percentage points over Stop-gradient, respectively.
In each backbone family, Full improves success on four tasks and matches Stop-gradient on the remaining task.
For the VLM-based model, the largest gains occur on Drawer and Box, at 15 and 10 percentage points; for the VGM-based model, the largest gain is on Cup, at 10 points.
C and D retain the same forward action conditioning: for identical parameters, inputs, and noise, $\operatorname{sg}(H_a)$ and $H_a$ yield identical forward predictions.
Their difference is whether future-loss gradients pass through the action-feature interface during training.
The resulting success-rate gains therefore support the value of this backward supervision for the deployed action-only policy.

\paragraph{The action-to-future connection adds to visual co-training (A, B, and D).}
Adding a context-only future objective increases average success from 83\% to 87\% for the VLM-based model and from 66\% to 72\% for the VGM-based model (B vs. A).
Full \model{} further improves over this control by 9 and 11 percentage points, respectively (D vs. B), with higher success on all five tasks in both backbone families.
Relative to action-only training, the total gains reach 13 points for the VLM-based model and 17 points for the VGM-based model (D vs. A).
Thus, visual co-training provides a useful learning signal, and connecting future prediction to action features yields additional gains.
D vs. B captures the combined contribution of forward conditioning and backward supervision, while D vs. C isolates the benefit of backpropagation through $H_a$.

\subsection{Q3: Does future-prediction error propagate to actions?}
\label{sec:future_error_sensitivity}
\label{sec:err_prop}
\label{sec:prediction}

We examine whether perturbing the predicted future of the same FastWAM-Joint evaluated for real-world task success in \Cref{sec:sim_results} changes its action error, and contrast this dependence with \model{}'s action-to-future design.
Appendix~\ref{sec:supp_wam_implementation} details the FastWAM-Joint implementation used here and in \Cref{sec:efficiency}.
We conduct open-loop evaluation across simulation and real-world tasks---RoboTwin, LIBERO, Drawer, Box, Shuttlecock, Cup, and Block---using the full validation set for each.
For each validation sample, we hold the observation, task instruction, and sampling randomness fixed across experimental conditions.

\paragraph{Perturbing FastWAM-Joint's future prediction.}
At an intermediate stage of joint denoising, we inject Gaussian noise once into the future-video latent, leaving the current observation and noised action chunk unchanged, and then resume normal sampling.
We evaluate perturbation strengths $\alpha\in\{0,0.1,0.25,0.5,1.0\}$, where $\alpha=0$ is the unperturbed condition, and report the percentage increase in action error relative to this condition.

\Needspace{0.36\textheight}
\begin{wrapfigure}{r}{0.5\textwidth}
    \ifdefined\capstart\capstart\fi
    \centering
    \includegraphics[width=\linewidth]{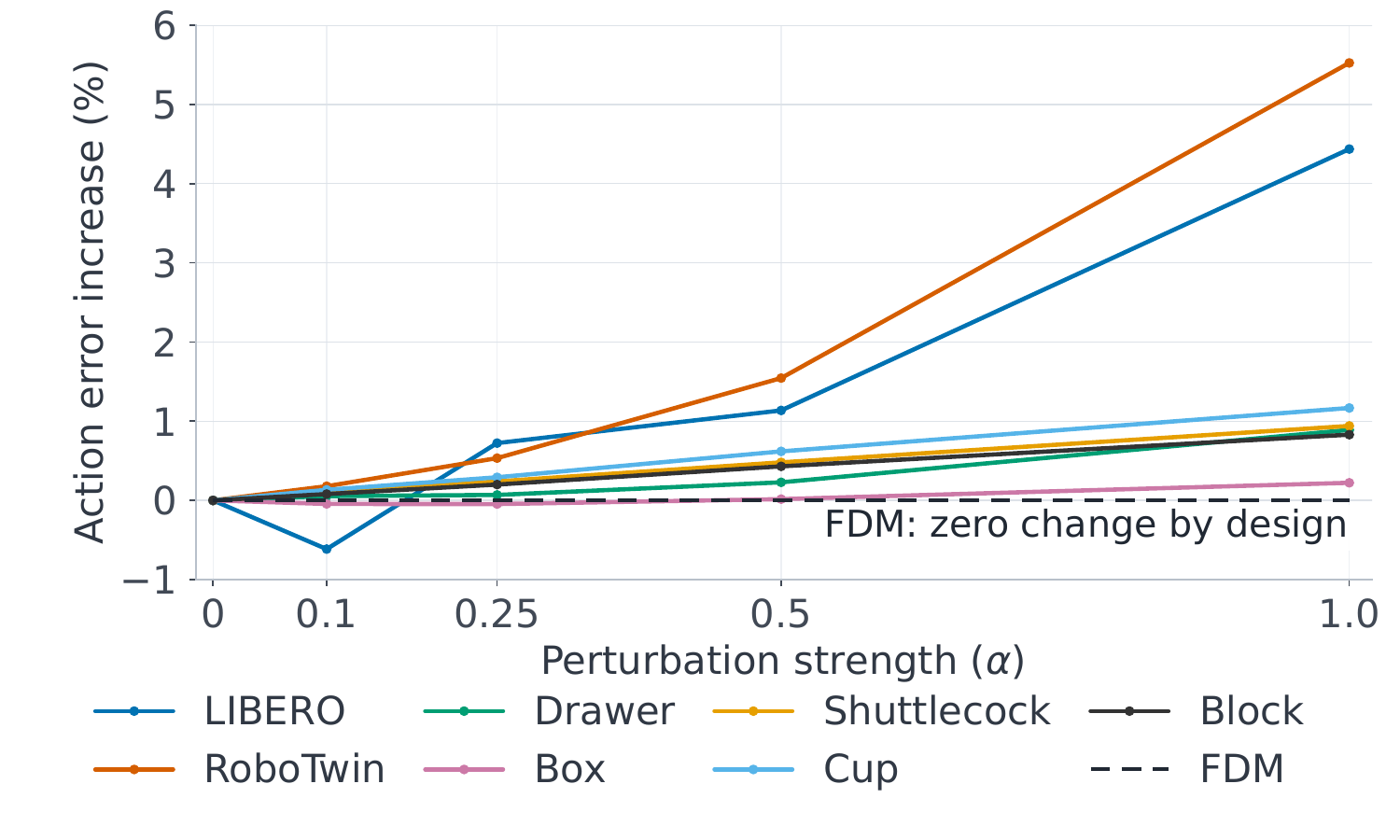}
    \vspace{-20pt}
    \caption{\textbf{Action sensitivity to future perturbations.}
    Solid curves show FastWAM-Joint's percentage increase in action error relative to $\alpha=0$.}
    \vspace{-5pt}
    \label{fig:wam_perturbation_action_error}
\end{wrapfigure}

\paragraph{Future perturbations affect FastWAM-Joint actions.}
\Cref{fig:wam_perturbation_action_error} shows that FastWAM-Joint's action error generally increases with perturbation strength across simulation and real-world tasks.
At $\alpha=1.0$, action error rises by 4.44\% on LIBERO and 5.52\% on RoboTwin; all five real-world tasks also show increases, ranging from 0.22\% on Box to 1.17\% on Cup.
Small perturbations yield slight decreases on LIBERO and Box, but all seven settings exhibit higher action error at $\alpha\geq0.5$, with the largest increase at $\alpha=1.0$.
These results show that perturbations to the generated future can affect FastWAM-Joint's action error even when the observation and noised action chunk are unchanged at the intervention step.

\begin{figure}[t]
    \centering
    \includegraphics[width=\linewidth]{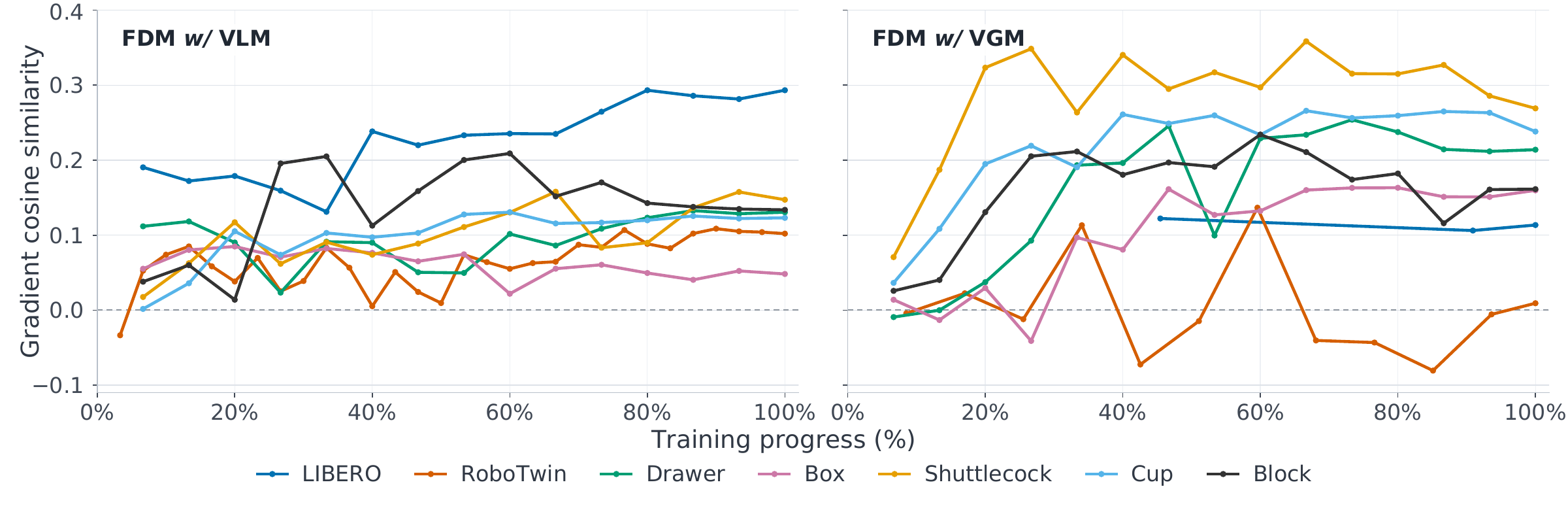}
    \caption{\textbf{Alignment between future and action gradients.}
    Cosine similarity between action-imitation gradients and future-loss gradients propagated through the action-to-future interface.}
    \label{fig:shared_kv_alignment}
    \vspace{-10pt}
\end{figure}

\paragraph{FDM's action inference is independent of future quality.}
In \model{}, the future-prediction branch reads action features, while the action path never reads generated future states.
Consequently, for fixed model parameters, inputs, and action-sampling randomness, perturbing the generated future leaves action predictions unchanged, regardless of future-generation quality.
The dashed zero line in \Cref{fig:wam_perturbation_action_error} denotes this architectural invariance: zero change in action error.
Thus, \model{} benefits from future supervision during training (\Cref{sec:ablation}) while keeping action inference independent of future-generation errors.

\subsection*{Q4: Are future and action gradients aligned?}
\label{sec:gradient_alignment}

We examine whether future supervision provides a locally compatible training signal for action imitation.
We compare action-imitation gradients with the future-loss gradient contribution propagated through the action-to-future interface, measuring their cosine similarity on shared-context key/value (K/V) projections.
This analysis isolates the backward contribution through action features, rather than using the total future-loss gradient.

\Cref{fig:shared_kv_alignment} reports gradient cosine similarity over training progress for \fdmvlm{} and \fdmvgm{}.
\fdmvlm{} exhibits predominantly positive alignment across the evaluated settings, with only limited early-stage deviations.
Most \fdmvgm{} settings also exhibit positive alignment after initial fluctuations.
These results indicate broadly compatible gradient directions between future supervision and action imitation in this analysis.
They support the use of the action-to-future connection to provide an auxiliary training signal for action learning, complementing the policy-level evidence in \Cref{sec:ablation}.

\subsection{Q5: What are the training cost and inference efficiency?}
\label{sec:efficiency}

We compare training step time and end-to-end action-chunk latency across FDM and the VLA and WAM paradigms within each backbone family under matched settings (\Cref{tab:efficiency}).
We use $\pi_{0.5}$ as the VLA method, while the evaluated future-conditioned FastWAM-Joint jointly supervises actions and futures during training and generates futures for action prediction at inference.
We evaluate the same architecture-matched FastWAM-Joint used for real-world task success in \Cref{sec:sim_results} and future perturbations in \Cref{sec:future_error_sensitivity}.
Future supervision increases training step time for the VLM-based policy while maintaining inference latency comparable to $\pi_{0.5}$.
For the VGM-based policy, training cost remains essentially unchanged, while action-only inference substantially reduces latency by removing future generation from the action prediction path.
Specifically, VLM-based training step time increases from 0.444 to 0.631~s ($+42.1\%$), while inference latency remains comparable at 202.98 versus 203.77~ms/chunk.
VGM-based training step time is nearly unchanged at 0.973 versus 0.975~s, while inference latency drops from 446.55 to 253.45~ms/chunk, yielding a $1.8\times$ speedup.

\begin{table}[htb]
    \caption{\textbf{Training cost and inference efficiency.}
    Comparisons use matched settings within each backbone family.
    (a) $\pi_{0.5}$ (VLA) uses action supervision; FastWAM-Joint uses joint action--future supervision.
    (b) Only FastWAM-Joint requires future generation for action prediction.
    FastWAM-Joint denotes the same future-conditioned method evaluated in \Cref{fig:real_future}.
    Relative metrics use $\pi_{0.5}$ for \fdmvlm{} and FastWAM-Joint for \fdmvgm{}.}
    \label{tab:efficiency}
    \begin{center}
    \scriptsize
    \setlength{\tabcolsep}{3pt}
    \renewcommand{\arraystretch}{1.05}
    \begin{minipage}[t]{0.52\linewidth}
        \centering
        \setlength{\tabcolsep}{1pt}
        \textbf{(a) Training cost}\par\vspace{3pt}
        \begin{tabular*}{\linewidth}{@{\extracolsep{\fill}}lcccc@{}}
            \toprule
            Method & Step time $\downarrow$ & Throughput $\uparrow$ & Memory $\downarrow$ & Overhead $\downarrow$ \\
            & (s/step) & (samples/s) & (GiB/GPU, peak) & (\%) \\
            \midrule
            $\pi_{0.5}$ & 0.4436 & 288.58 & 67.08 & --- \\
            \fdmvlm{} & 0.6305 & 203.01 & 72.04 & $+42.14$ \\
            \midrule
            FastWAM-Joint & 0.9754 & 131.23 & 66.80 & --- \\
            \fdmvgm{} & 0.9729 & 131.56 & 66.80 & $-0.26$ \\
            \bottomrule
        \end{tabular*}
    \end{minipage}\hfill
    \begin{minipage}[t]{0.47\linewidth}
        \centering
        \setlength{\tabcolsep}{1pt}
        \textbf{(b) Inference efficiency}\par\vspace{3pt}
        \begin{tabular*}{\linewidth}{@{\extracolsep{\fill}}lccc@{}}
            \toprule
            Method & Latency $\downarrow$ & Speedup $\uparrow$ & Reduction $\uparrow$ \\
            & (ms/chunk) & ($\times$) & (\%) \\
            \midrule
            $\pi_{0.5}$ & 203.77 & $1.00$ & $0.00$ \\
            \fdmvlm{} & 202.98 & $1.00$ & $0.39$ \\
            \midrule
            FastWAM-Joint & 446.55 & $1.00$ & $0.00$ \\
            \fdmvgm{} & 253.45 & $1.76$ & $43.24$ \\
            \bottomrule
        \end{tabular*}
    \end{minipage}
    \end{center}
\end{table}
\vspace{-10pt}

\section{Conclusion}

We introduced FDM, a family of robotic action models built around action-to-future learning.
Actions and future world states are jointly predicted with asymmetric conditioning: future prediction reads action features, while action prediction remains independent of predicted futures.
Future prediction errors supervise these action features through backpropagation.
Ground-truth future states thus supervise the policy during training, while action-only inference skips future generation.
Our VLM- and VGM-based instantiations realize this principle via future-image latent regression and future-video flow matching, respectively.
They achieve strong LIBERO performance and improve success on RoboTwin and five real-world tasks.
Controlled ablations and gradient analysis examine backward future supervision, while future-latent perturbations test action sensitivity to prediction errors at inference.
\fdmvlm{} retains comparable action latency, while \fdmvgm{} reduces latency relative to FastWAM-Joint.

\FloatBarrier

\section*{AI use statement}

Generative AI tools were used only to assist with writing, LaTeX formatting, and information retrieval during manuscript preparation.
They were not used to develop the method, design experiments, implement research code, process data, or interpret results.
The authors manually reviewed the AI-assisted content and checked the references.
The authors take responsibility for the final content of this work, including any text, claims, or artifacts prepared with the assistance of generative AI.

\section*{Reproducibility statement}

To support reproducibility, \Cref{sec:method} describes the action--future information flow, the VLM- and VGM-based instantiations, their training objectives, and action-only inference.
\Cref{sec:setup} identifies the backbone families, comparison methods, and evaluation benchmarks, while \Cref{sec:ablation} describes the supervision variants and matched comparison design.
Appendix~\ref{sec:supp_real_tasks} documents the real-robot hardware, camera configuration, teleoperation interface, and task descriptions.
Appendix~\ref{sec:supp_hyperparameters} reports the training hyperparameters for \fdmvlm{} and \fdmvgm{} used in the real-world experiments, including optimization settings, batch sizes, input resolutions, and action prediction horizons.

\bibliographystyle{iclr2027_conference}
\bibliography{main}

\begin{thebibliography}{67}
\providecommand{\natexlab}[1]{#1}
\providecommand{\url}[1]{\texttt{#1}}
\expandafter\ifx\csname urlstyle\endcsname\relax
  \providecommand{\doi}[1]{doi: #1}\else
  \providecommand{\doi}{doi: \begingroup \urlstyle{rm}\Url}\fi

\bibitem[{AgiBot Research Team} et~al.(2026){AgiBot Research Team}, Liu, Zhao,
  Yang, Chen, Zhou, Chen, Ren, Peng, Jin, Wang, Wang, He, Feng, Xiong, Zhong,
  Wei, Han, Zhang, Huang, Zhao, Yin, Wu, Yan, Hu, Yan, Niyazi, Fang, Zeng,
  Meng, Lv, Cao, Hou, Ye, Niu, Cai, Hu, Min, Cai, Liao, Wu, Poria, Li, Zhou,
  and Yao]{agibot2026geact2}
{AgiBot Research Team}, Renhang Liu, Wenzhi Zhao, Zhuo Yang, Liliang Chen,
  Pengfei Zhou, Shengcong Chen, Guanghui Ren, Youlun Peng, Rongjun Jin, Nan
  Wang, Sukai Wang, Xindong He, Jinyuan Feng, Ziyu Xiong, Linqing Zhong, Yifei
  Wei, Feng Han, Long Zhang, Da~Huang, Nanshu Zhao, Chenghao Yin, Mo~Wu,
  Zhaodong Yan, Kongtao Hu, Yuxiang Yan, Aogelijiang Niyazi, Yu~Fang, Jia Zeng,
  Lizhu Meng, Daizhen Lv, Haoyu Cao, Zhiwen Hou, Lianjin Ye, Yuehan Niu, Zhikai
  Cai, Xuan Hu, Hui Min, Xiongfeng Cai, Yue Liao, Jing Wu, Soujanya Poria,
  Ye~Li, Sanping Zhou, and Maoqing Yao.
\newblock Ge-act 2.0: Pretraining and scaling a world-action model for robotic
  manipulation, 2026.

\bibitem[Agrawal et~al.(2016)Agrawal, Nair, Abbeel, Malik, and
  Levine]{agrawal2016poke}
Pulkit Agrawal, Ashvin Nair, Pieter Abbeel, Jitendra Malik, and Sergey Levine.
\newblock Learning to poke by poking: Experiential learning of intuitive
  physics.
\newblock In \emph{Advances in Neural Information Processing Systems
  (NeurIPS)}, 2016.

\bibitem[Ali et~al.(2025)Ali, Bai, Bala, Balaji, Blakeman, Cai, Cao, Cao, Cha,
  Chao, et~al.]{ali2025world}
Arslan Ali, Junjie Bai, Maciej Bala, Yogesh Balaji, Aaron Blakeman, Tiffany
  Cai, Jiaxin Cao, Tianshi Cao, Elizabeth Cha, Yu-Wei Chao, et~al.
\newblock World simulation with video foundation models for physical ai.
\newblock \emph{arXiv preprint}, 2025.

\bibitem[Bi et~al.(2026{\natexlab{a}})Bi, Tan, Xie, Wang, Huang, Liu, Zhao,
  Feng, Xiang, Rong, et~al.]{bi2026motus}
Hongzhe Bi, Hengkai Tan, Shenghao Xie, Zeyuan Wang, Shuhe Huang, Haitian Liu,
  Ruowen Zhao, Yao Feng, Chendong Xiang, Yinze Rong, et~al.
\newblock Motus: A unified latent action world model.
\newblock In \emph{CVPR}, 2026{\natexlab{a}}.

\bibitem[Bi et~al.(2026{\natexlab{b}})Bi, Zhou, Tang, Pang, Huang, Liu, Wang,
  Huang, Wang, Cheng, et~al.]{bi2026motus2}
Hongzhe Bi, Zihao Zhou, Yihang Tang, Jingrui Pang, Shuhe Huang, Haitian Liu,
  Runqing Wang, Shuai Huang, Yichen Wang, Yiming Cheng, et~al.
\newblock Motus2: A self-evolving general world model for dexterous
  manipulation.
\newblock \emph{arXiv preprint}, 2026{\natexlab{b}}.

\bibitem[Black et~al.(2024)Black, Brown, Driess, Esmail, Equi, Finn, Fusai,
  Groom, Hausman, Ichter, et~al.]{black2024pi_0}
Kevin Black, Noah Brown, Danny Driess, Adnan Esmail, Michael Equi, Chelsea
  Finn, Niccolo Fusai, Lachy Groom, Karol Hausman, Brian Ichter, et~al.
\newblock {$\pi_0$}: A vision-language-action flow model for general robot
  control.
\newblock \emph{arXiv preprint}, 2024.

\bibitem[Cai et~al.(2026{\natexlab{a}})Cai, Ling, Chu, Liu, Kang, Liang, Xu,
  Mao, Zhang, Yang, et~al.]{cai2026aha}
Jisong Cai, Long Ling, Shiwei Chu, Zhongshan Liu, Jiayue Kang, Zhixuan Liang,
  Wenjie Xu, Yinan Mao, Weinan Zhang, Xiaokang Yang, et~al.
\newblock Aha-wam: Asynchronous horizon-adaptive world-action modeling with
  observation-guided context routing.
\newblock \emph{arXiv preprint}, 2026{\natexlab{a}}.

\bibitem[Cai et~al.(2026{\natexlab{b}})Cai, Cai, Cao, Chen, He, Jiang, Li, Li,
  Li, Liu, et~al.]{internvla_a1}
Junhao Cai, Zetao Cai, Jiafei Cao, Yilun Chen, Zeyu He, Lei Jiang, Hang Li,
  Hengjie Li, Yang Li, Yufei Liu, et~al.
\newblock Internvla-a1: Unifying understanding, generation and action for
  robotic manipulation.
\newblock \emph{arXiv preprint}, 2026{\natexlab{b}}.

\bibitem[Cen et~al.(2025{\natexlab{a}})Cen, Huang, Yuan, Li, Yuan, Yu, Hou,
  Jiang, Guo, Li, et~al.]{cen2025rynnvla}
Jun Cen, Siteng Huang, Yuqian Yuan, Kehan Li, Hangjie Yuan, Chaohui Yu, Bohan
  Hou, Yuming Jiang, Jiayan Guo, Xin Li, et~al.
\newblock Rynnvla-002: A unified vision-language-action and world model.
\newblock \emph{arXiv preprint}, 2025{\natexlab{a}}.

\bibitem[Cen et~al.(2025{\natexlab{b}})Cen, Yu, Yuan, Jiang, Huang, Guo, Li,
  Song, Luo, Wang, et~al.]{cen2025worldvla}
Jun Cen, Chaohui Yu, Hangjie Yuan, Yuming Jiang, Siteng Huang, Jiayan Guo, Xin
  Li, Yibing Song, Hao Luo, Fan Wang, et~al.
\newblock Worldvla: Towards autoregressive action world model.
\newblock \emph{arXiv preprint}, 2025{\natexlab{b}}.

\bibitem[Cheang et~al.(2024)Cheang, Chen, Jing, Kong, Li, Li, Liu, Wu, Xu,
  Yang, et~al.]{cheang2024gr}
Chi-Lam Cheang, Guangzeng Chen, Ya~Jing, Tao Kong, Hang Li, Yifeng Li, Yuxiao
  Liu, Hongtao Wu, Jiafeng Xu, Yichu Yang, et~al.
\newblock Gr-2: A generative video-language-action model with web-scale
  knowledge for robot manipulation.
\newblock \emph{arXiv preprint}, 2024.

\bibitem[Chen et~al.(2026)Chen, Jia, Wuwu, Liu, Du, Fan, Chi, Chen, Bai, Qian,
  et~al.]{chen2026mvwam}
Jintao Chen, Peidong Jia, Qingpo Wuwu, Jiaming Liu, Mengfei Du, Chun-Kai Fan,
  Xiaowei Chi, Hao Chen, Chengyu Bai, Zezhong Qian, et~al.
\newblock Mv-wam: Manifold-aware world action model with value augmentation.
\newblock \emph{arXiv preprint}, 2026.

\bibitem[Chen et~al.(2025)Chen, Chen, Chen, Cai, Liu, Li, Liang, Lin, Ge, Gu,
  et~al.]{chen2025robotwin}
Tianxing Chen, Zanxin Chen, Baijun Chen, Zijian Cai, Yibin Liu, Zixuan Li,
  Qiwei Liang, Xianliang Lin, Yiheng Ge, Zhenyu Gu, et~al.
\newblock Robotwin 2.0: A scalable data generator and benchmark with strong
  domain randomization for robust bimanual robotic manipulation.
\newblock \emph{arXiv preprint}, 2025.

\bibitem[Chi et~al.(2025)Chi, Xu, Feng, Cousineau, Du, Burchfiel, Tedrake, and
  Song]{Diffusionpolicy}
Cheng Chi, Zhenjia Xu, Siyuan Feng, Eric Cousineau, Yilun Du, Benjamin
  Burchfiel, Russ Tedrake, and Shuran Song.
\newblock Diffusion policy: Visuomotor policy learning via action diffusion.
\newblock \emph{The International Journal of Robotics Research}, 2025.

\bibitem[Contributors(2025)]{internvlam1}
InternVLA-M1 Contributors.
\newblock Internvla-m1: A spatially guided vision-language-action framework for
  generalist robot policy.
\newblock \emph{arXiv preprint}, 2025.

\bibitem[Feng et~al.(2025)Feng, Tan, Mao, Xiang, Liu, Huang, Su, and
  Zhu]{feng2025vidar}
Yao Feng, Hengkai Tan, Xinyi Mao, Chendong Xiang, Guodong Liu, Shuhe Huang,
  Hang Su, and Jun Zhu.
\newblock Vidar: Embodied video diffusion model for generalist manipulation.
\newblock \emph{arXiv preprint}, 2025.

\bibitem[Guo et~al.(2026)Guo, Li, Li, Chen, Sun, Su, Wang, Zhang, Li, and
  Liu]{guo2026xwam}
Jun Guo, Qiwei Li, Peiyan Li, Zilong Chen, Nan Sun, Yifei Su, Heyun Wang, Yuan
  Zhang, Xinghang Li, and Huaping Liu.
\newblock Unified 4d world action modeling from video priors with asynchronous
  denoising.
\newblock \emph{arXiv preprint}, 2026.

\bibitem[HaCohen et~al.(2024)HaCohen, Chiprut, Brazowski, Shalem, Moshe,
  Richardson, Levin, Shiran, Zabari, Gordon, et~al.]{hacohen2024ltx}
Yoav HaCohen, Nisan Chiprut, Benny Brazowski, Daniel Shalem, Dudu Moshe, Eitan
  Richardson, Eran Levin, Guy Shiran, Nir Zabari, Ori Gordon, et~al.
\newblock Ltx-video: Realtime video latent diffusion.
\newblock \emph{arXiv preprint}, 2024.

\bibitem[Hu et~al.(2024)Hu, Guo, Wang, Chen, Wang, Zhang, Sreenath, Lu, and
  Chen]{hu2024video}
Yucheng Hu, Yanjiang Guo, Pengchao Wang, Xiaoyu Chen, Yen-Jen Wang, Jianke
  Zhang, Koushil Sreenath, Chaochao Lu, and Jianyu Chen.
\newblock Video prediction policy: A generalist robot policy with predictive
  visual representations.
\newblock \emph{arXiv preprint}, 2024.

\bibitem[Hu et~al.(2026)Hu, Zhang, Luo, Guo, Chen, Sun, Feng, Lu, Chen, Zhang,
  et~al.]{hu2026bagelvla}
Yucheng Hu, Jianke Zhang, Yuanfei Luo, Yanjiang Guo, Xiaoyu Chen, Xinshu Sun,
  Kun Feng, Qingzhou Lu, Sheng Chen, Yangang Zhang, et~al.
\newblock Bagelvla: Enhancing long-horizon manipulation via interleaved
  vision-language-action generation.
\newblock \emph{arXiv preprint}, 2026.

\bibitem[Intelligence et~al.(2025)Intelligence, Black, Brown, Darpinian,
  Dhabalia, Driess, Esmail, Equi, Finn, Fusai, et~al.]{intelligence2025pi_}
Physical Intelligence, Kevin Black, Noah Brown, James Darpinian, Karan
  Dhabalia, Danny Driess, Adnan Esmail, Michael Equi, Chelsea Finn, Niccolo
  Fusai, et~al.
\newblock {$\pi_{0.5}$}: a vision-language-action model with open-world
  generalization.
\newblock \emph{arXiv preprint}, 2025.

\bibitem[Jordan \& Rumelhart(1992)Jordan and Rumelhart]{jordan1992forward}
Michael~I. Jordan and David~E. Rumelhart.
\newblock Forward models: Supervised learning with a distal teacher.
\newblock \emph{Cognitive Science}, 16\penalty0 (3):\penalty0 307--354, 1992.

\bibitem[Kang et~al.(2026)Kang, Shi, Liang, Gan, Liu, Zhang, Chen, Qin, Zheng,
  Zheng, et~al.]{kang2026xtokenizer}
Miracle Kang, Lights Shi, Lucy Liang, Roy Gan, Dongxiu Liu, Pushi Zhang, Sylas
  Chen, Shawn Qin, Yinan Zheng, Jinliang Zheng, et~al.
\newblock X-tokenizer: A multimodal action tokenizer for vision-language-action
  pretraining.
\newblock \emph{arXiv preprint}, 2026.

\bibitem[Kim et~al.(2024)Kim, Pertsch, Karamcheti, Xiao, Balakrishna, Nair,
  Rafailov, Foster, Lam, Sanketi, et~al.]{kim24openvla}
Moo~Jin Kim, Karl Pertsch, Siddharth Karamcheti, Ted Xiao, Ashwin Balakrishna,
  Suraj Nair, Rafael Rafailov, Ethan Foster, Grace Lam, Pannag Sanketi, et~al.
\newblock Openvla: An open-source vision-language-action model.
\newblock \emph{arXiv preprint}, 2024.

\bibitem[Kim et~al.(2026)Kim, Gao, Lin, Lin, Ge, Lam, Liang, Song, Liu, Finn,
  and Gu]{kim2026cosmos}
Moo~Jin Kim, Yihuai Gao, Tsung-Yi Lin, Yen-Chen Lin, Yunhao Ge, Grace Lam,
  Percy Liang, Shuran Song, Ming-Yu Liu, Chelsea Finn, and Jinwei Gu.
\newblock Cosmos policy: Fine-tuning video models for visuomotor control and
  planning.
\newblock \emph{arXiv preprint}, 2026.

\bibitem[Li et~al.(2026{\natexlab{a}})Li, Zhang, Luo, Yang, Wang, Han, Yu, Gao,
  Xue, Zhu, Shen, and Xu]{lingbot-va2026}
Lin Li, Qihang Zhang, Yiming Luo, Shuai Yang, Ruilin Wang, Fei Han, Mingrui Yu,
  Zelin Gao, Nan Xue, Xing Zhu, Yujun Shen, and Yinghao Xu.
\newblock Causal world modeling for robot control.
\newblock \emph{arXiv preprint}, 2026{\natexlab{a}}.

\bibitem[Li et~al.(2026{\natexlab{b}})Li, Yao, Yang, Qu, Cheng, Yu, Lu, Von,
  Chen, Tang, et~al.]{li2026wall}
Shalfun Li, Victor Yao, Charles Yang, Truth Qu, Regis Cheng, Ryan Yu, Howard
  Lu, Newton Von, Vincent Chen, Yohann Tang, et~al.
\newblock Wall-wm: Carving world action modeling at the event joints.
\newblock \emph{arXiv preprint}, 2026{\natexlab{b}}.

\bibitem[Li et~al.(2025)Li, Gao, Sadigh, and Song]{li2503unified}
Shuang Li, Yihuai Gao, Dorsa Sadigh, and Shuran Song.
\newblock Unified video action model, 2025.
\newblock \emph{arXiv preprint}, 2025.

\bibitem[Liang et~al.(2024)Liang, Yu, Luo, Iyer, Dong, Zhou, Ghosh, Lewis, Yih,
  Zettlemoyer, et~al.]{liang2024mixture}
Weixin Liang, Lili Yu, Liang Luo, Srinivasan Iyer, Ning Dong, Chunting Zhou,
  Gargi Ghosh, Mike Lewis, Wen-tau Yih, Luke Zettlemoyer, et~al.
\newblock Mixture-of-transformers: A sparse and scalable architecture for
  multi-modal foundation models.
\newblock \emph{arXiv preprint}, 2024.

\bibitem[Liu et~al.(2023)Liu, Zhu, Gao, Feng, Liu, Zhu, and
  Stone]{liu2023libero}
Bo~Liu, Yifeng Zhu, Chongkai Gao, Yihao Feng, Qiang Liu, Yuke Zhu, and Peter
  Stone.
\newblock Libero: Benchmarking knowledge transfer for lifelong robot learning.
\newblock \emph{arXiv preprint}, 2023.

\bibitem[Liu et~al.(2026)Liu, Hou, Lu, Yang, Fan, Luo, Li, Han, Hou, and
  Zhao]{liu2026streampi}
Zhe Liu, Jinghua Hou, Yuxiang Lu, Zhenya Yang, Xianzhe Fan, Junwei Luo, Junyi
  Li, Ruihua Han, Zhi Hou, and Hengshuang Zhao.
\newblock Streampi: Streaming multimodal temporal modeling for
  vision-language-action models.
\newblock \emph{arXiv preprint}, 2026.

\bibitem[Lv et~al.(2025)Lv, Kong, Li, Zeng, Qiu, Qu, Song, Chen, Deng, and
  Pang]{lv2025f1}
Qi~Lv, Weijie Kong, Hao Li, Jia Zeng, Zherui Qiu, Delin Qu, Haoming Song, Qizhi
  Chen, Xiang Deng, and Jiangmiao Pang.
\newblock F1: A vision-language-action model bridging understanding and
  generation to actions.
\newblock \emph{arXiv preprint}, 2025.

\bibitem[Ma et~al.(2026{\natexlab{a}})Ma, Peng, Yue, Cao, Wang, Zhang, and
  Ma]{ma2026geosem}
Fulong Ma, Daojie Peng, Wenjun Yue, Jiahang Cao, Bintao Wang, Qiang Zhang, and
  Jun Ma.
\newblock Geosem-wam: Geometry- and semantic-aware world action models.
\newblock \emph{arXiv preprint}, 2026{\natexlab{a}}.

\bibitem[Ma et~al.(2026{\natexlab{b}})Ma, Cai, Xu, Li, Yang, Tian, Cao, Zhu,
  Qiu, Yang, et~al.]{ma2026internvla}
Haoxiang Ma, Junhao Cai, Xiaoxu Xu, Hao Li, Yuyin Yang, Yang Tian, Jiafei Cao,
  Hongrui Zhu, Zherui Qiu, Yuqiang Yang, et~al.
\newblock Internvla-a1.5: Unifying understanding, latent foresight, and action
  for compositional generalization.
\newblock \emph{arXiv preprint}, 2026{\natexlab{b}}.

\bibitem[Ma et~al.(2026{\natexlab{c}})Ma, Zheng, Wang, Jiang, Cui, Liang, and
  Yang]{ma2026dit4dit}
Teli Ma, Jia Zheng, Zifan Wang, Chunli Jiang, Andy Cui, Junwei Liang, and Shuo
  Yang.
\newblock Dit4dit: Jointly modeling video dynamics and actions for
  generalizable robot control.
\newblock \emph{arXiv preprint}, 2026{\natexlab{c}}.

\bibitem[NVIDIA et~al.(2025)NVIDIA, Bjorck, Castañeda, Cherniadev, Da, Ding,
  Fan, Fang, Fox, Hu, Huang, Jang, Jiang, Kautz, Kundalia, Lao, Li, Lin, Lin,
  Liu, Llontop, Magne, Mandlekar, Narayan, Nasiriany, Reed, Tan, Wang, Wang,
  Wang, Wang, Xiang, Xie, Xu, Xu, Ye, Yu, Zhang, Zhang, Zhao, Zheng, and
  Zhu]{GR00T}
NVIDIA, Johan Bjorck, Fernando Castañeda, Nikita Cherniadev, Xingye Da, Runyu
  Ding, Linxi~"Jim" Fan, Yu~Fang, Dieter Fox, Fengyuan Hu, Spencer Huang, Joel
  Jang, Zhenyu Jiang, Jan Kautz, Kaushil Kundalia, Lawrence Lao, Zhiqi Li,
  Zongyu Lin, Kevin Lin, Guilin Liu, Edith Llontop, Loic Magne, Ajay Mandlekar,
  Avnish Narayan, Soroush Nasiriany, Scott Reed, You~Liang Tan, Guanzhi Wang,
  Zu~Wang, Jing Wang, Qi~Wang, Jiannan Xiang, Yuqi Xie, Yinzhen Xu, Zhenjia Xu,
  Seonghyeon Ye, Zhiding Yu, Ao~Zhang, Hao Zhang, Yizhou Zhao, Ruijie Zheng,
  and Yuke Zhu.
\newblock Gr00t n1: An open foundation model for generalist humanoid robots,
  2025.

\bibitem[Pai et~al.(2025)Pai, Achenbach, Montesinos, Forrai, Mees, and
  Nava]{pai2025mimic}
Jonas Pai, Liam Achenbach, Victoriano Montesinos, Benedek Forrai, Oier Mees,
  and Elvis Nava.
\newblock mimic-video: Video-action models for generalizable robot control
  beyond vlas.
\newblock \emph{arXiv preprint}, 2025.

\bibitem[Qu et~al.(2025)Qu, Song, Chen, Yao, Ye, Ding, Wang, Gu, Zhao, Wang,
  et~al.]{qu2025spatialvla}
Delin Qu, Haoming Song, Qizhi Chen, Yuanqi Yao, Xinyi Ye, Yan Ding, Zhigang
  Wang, JiaYuan Gu, Bin Zhao, Dong Wang, et~al.
\newblock Spatialvla: Exploring spatial representations for
  visual-language-action model.
\newblock \emph{RSS}, 2025.

\bibitem[Shukor et~al.(2025)Shukor, Aubakirova, Capuano, Kooijmans, Palma,
  Zouitine, Aractingi, Pascal, Russi, Marafioti, et~al.]{shukor2025smolvla}
Mustafa Shukor, Dana Aubakirova, Francesco Capuano, Pepijn Kooijmans, Steven
  Palma, Adil Zouitine, Michel Aractingi, Caroline Pascal, Martino Russi,
  Andres Marafioti, et~al.
\newblock Smolvla: A vision-language-action model for affordable and efficient
  robotics.
\newblock \emph{arXiv preprint}, 2025.

\bibitem[Tian et~al.(2026)Tian, Zheng, Zheng, Gu, Zang, Qin, Li, Li, Ding, and
  Zhao]{tian2026vtwam}
Shuai Tian, Yupeng Zheng, Yuhang Zheng, Songen Gu, Yujie Zang, Yuxing Qin,
  Weize Li, Haoran Li, Wenchao Ding, and Dongbin Zhao.
\newblock Vt-wam: Visual-tactile world action model for contact-rich
  manipulation.
\newblock \emph{arXiv preprint}, 2026.

\bibitem[Tian et~al.(2024)Tian, Yang, Zeng, Wang, Lin, Dong, and
  Pang]{tian2024seer}
Yang Tian, Sizhe Yang, Jia Zeng, Ping Wang, Dahua Lin, Hao Dong, and Jiangmiao
  Pang.
\newblock Predictive inverse dynamics models are scalable learners for robotic
  manipulation.
\newblock \emph{arXiv preprint}, 2024.

\bibitem[Wan et~al.(2025)Wan, Wang, Ai, Wen, Mao, Xie, Chen, Yu, Zhao, Yang,
  et~al.]{wan2025wan}
Team Wan, Ang Wang, Baole Ai, Bin Wen, Chaojie Mao, Chen-Wei Xie, Di~Chen,
  Feiwu Yu, Haiming Zhao, Jianxiao Yang, et~al.
\newblock Wan: Open and advanced large-scale video generative models.
\newblock \emph{arXiv preprint}, 2025.

\bibitem[Wang et~al.(2026{\natexlab{a}})Wang, Li, Guan, Ye, Xie, Liu, Chen,
  Liang, Zhang, Hu, et~al.]{wang2026qwen}
Qiuyue Wang, Mingsheng Li, Jian Guan, Jinhui Ye, Sicheng Xie, Yitao Liu, Junhao
  Chen, Zhixuan Liang, Jie Zhang, Xintong Hu, et~al.
\newblock Qwen-vla: Unifying vision-language-action modeling across tasks,
  environments, and robot embodiments.
\newblock \emph{arXiv preprint}, 2026{\natexlab{a}}.

\bibitem[Wang et~al.(2025)Wang, Ding, Li, Cui, Ge, Tong, Song, Zhao, Zhao, Hou,
  et~al.]{wang2025vlaadapter}
Yihao Wang, Pengxiang Ding, Lingxiao Li, Can Cui, Zirui Ge, Xinyang Tong,
  Wenxuan Song, Han Zhao, Wei Zhao, Pengxu Hou, et~al.
\newblock Vla-adapter: An effective paradigm for tiny-scale
  vision-language-action model.
\newblock \emph{arXiv preprint}, 2025.

\bibitem[Wang et~al.(2026{\natexlab{b}})Wang, Li, Wang, Zhang, Li, Chen, Wang,
  and Zhang]{wang2025unified}
Yuqi Wang, Xinghang Li, Wenxuan Wang, Junbo Zhang, Yingyan Li, Yuntao Chen,
  Xinlong Wang, and Zhaoxiang Zhang.
\newblock Unified vision-language-action model.
\newblock In \emph{ICLR}, 2026{\natexlab{b}}.

\bibitem[Wang et~al.(2026{\natexlab{c}})Wang, Huang, Li, Zhang, Liang, Jin,
  Chen, Chi, Zhou, Yu, et~al.]{wang2026openwam}
Yuran Wang, Siqiao Huang, Mingleyang Li, Chenhao Zhang, Jiaqi Liang, Weiyang
  Jin, Yue Chen, Xuemin Chi, Donghao Zhou, Qize Yu, et~al.
\newblock Openwam: An open, modular exploration towards systematic world-action
  model pretraining.
\newblock \emph{arXiv preprint}, 2026{\natexlab{c}}.

\bibitem[Won et~al.(2025)Won, Lee, Jang, Kim, and Shin]{won2025dust}
John Won, Kyungmin Lee, Huiwon Jang, Dongyoung Kim, and Jinwoo Shin.
\newblock Dual-stream diffusion for world-model augmented
  vision-language-action model.
\newblock \emph{arXiv preprint}, 2025.

\bibitem[Wu et~al.(2023)Wu, Jing, Cheang, Chen, Xu, Li, Liu, Li, and
  Kong]{wu2023gr1}
Hongtao Wu, Ya~Jing, Chilam Cheang, Guangzeng Chen, Jiafeng Xu, Xinghang Li,
  Minghuan Liu, Hang Li, and Tao Kong.
\newblock Unleashing large-scale video generative pre-training for visual robot
  manipulation.
\newblock \emph{arXiv preprint}, 2023.

\bibitem[Xu et~al.(2026)Xu, Zhang, Zhou, Zhu, Shen, Yang, and
  Xu]{xu2026nextforcing}
Gangwei Xu, Qihang Zhang, Jiaming Zhou, Xing Zhu, Yujun Shen, Xin Yang, and
  Yinghao Xu.
\newblock Next forcing: Causal world modeling with multi-chunk prediction.
\newblock \emph{arXiv preprint}, 2026.

\bibitem[Yang et~al.(2026)Yang, Zeng, Lin, Chang, Qi, Xiao, Liu, Chen, Chen,
  Huo, et~al.]{yang2026abot}
Yandan Yang, Shuang Zeng, Tong Lin, Xinyuan Chang, Dekang Qi, Junjin Xiao,
  Haoyun Liu, Ronghan Chen, Yuzhi Chen, Dongjie Huo, et~al.
\newblock Abot-m0: Vla foundation model for robotic manipulation with action
  manifold learning.
\newblock \emph{arXiv preprint}, 2026.

\bibitem[Yang et~al.(2025)Yang, Li, Chen, Song, Wang, Xiao, Su, Qiaoben, Liu,
  and Deng]{yang2025mantis}
Yi~Yang, Xueqi Li, Yiyang Chen, Jin Song, Yihan Wang, Zipeng Xiao, Jiadi Su,
  You Qiaoben, Pengfei Liu, and Zhijie Deng.
\newblock Mantis: A versatile vision-language-action model with disentangled
  visual foresight.
\newblock \emph{arXiv preprint}, 2025.

\bibitem[Ye et~al.(2026{\natexlab{a}})Ye, Wang, Ni, Huang, Zhao, Li, Li, Li,
  Lv, Liu, et~al.]{ye2026gigaworld}
Angen Ye, Boyuan Wang, Chaojun Ni, Guan Huang, Guosheng Zhao, Hao Li, Hengtao
  Li, Jie Li, Jindi Lv, Jingyu Liu, et~al.
\newblock Gigaworld-policy: An efficient action-centered world--action model.
\newblock \emph{arXiv preprint}, 2026{\natexlab{a}}.

\bibitem[Ye et~al.(2026{\natexlab{b}})Ye, Gao, Yang, Zheng, Wang, Chen, Chen,
  Chen, Liu, and Jia]{ye2026starvla}
Jinhui Ye, Ning Gao, Senqiao Yang, Jinliang Zheng, Zixuan Wang, Yuxin Chen,
  Pengguang Chen, Yilun Chen, Shu Liu, and Jiaya Jia.
\newblock Starvla-{$\alpha$}: Reducing complexity in vision-language-action
  systems.
\newblock \emph{arXiv preprint}, 2026{\natexlab{b}}.

\bibitem[Ye et~al.(2026{\natexlab{c}})Ye, Ge, Zheng, Gao, Yu, Kurian, Indupuru,
  Tan, Zhu, Xiang, Malik, Lee, Liang, Ranawaka, Gu, Xu, Wang, Hu, Narayan,
  Bjorck, Wang, Kim, Niu, Zheng, Xie, Wu, Wang, Julian, Xu, Du, Chebotar, Reed,
  Kautz, Zhu, Fan, and Jang]{ye2026worldactionmodelszeroshot}
Seonghyeon Ye, Yunhao Ge, Kaiyuan Zheng, Shenyuan Gao, Sihyun Yu, George
  Kurian, Suneel Indupuru, You~Liang Tan, Chuning Zhu, Jiannan Xiang, Ayaan
  Malik, Kyungmin Lee, William Liang, Nadun Ranawaka, Jiasheng Gu, Yinzhen Xu,
  Guanzhi Wang, Fengyuan Hu, Avnish Narayan, Johan Bjorck, Jing Wang, Gwanghyun
  Kim, Dantong Niu, Ruijie Zheng, Yuqi Xie, Jimmy Wu, Qi~Wang, Ryan Julian,
  Danfei Xu, Yilun Du, Yevgen Chebotar, Scott Reed, Jan Kautz, Yuke Zhu,
  Linxi~"Jim" Fan, and Joel Jang.
\newblock World action models are zero-shot policies, 2026{\natexlab{c}}.

\bibitem[Yu et~al.(2026{\natexlab{a}})Yu, You, Wang, Liang, Ping, Tian, Chen,
  Cai, Gong, Wu, et~al.]{yu2026affordancevla}
Qize Yu, Jiadi You, Yuran Wang, Jiaqi Liang, Bowen Ping, Yang Tian, Yue Chen,
  Minghong Cai, Zeying Gong, Ruihai Wu, et~al.
\newblock Affordancevla: A vision-language-action model empowering action
  generation through affordance-aware understanding.
\newblock \emph{arXiv preprint}, 2026{\natexlab{a}}.

\bibitem[Yu et~al.(2026{\natexlab{b}})Yu, Zhang, Liu, Liu, Kang, Li, Shi, Ma,
  Yang, Pan, et~al.]{yu2026wall}
Ryan Yu, Pushi Zhang, Starrick Liu, Brae Liu, Miracle Kang, Shalfun Li, Lights
  Shi, Ellie Ma, Ping Yang, Chris Pan, et~al.
\newblock Wall-oss-0.5 technical report.
\newblock \emph{arXiv preprint}, 2026{\natexlab{b}}.

\bibitem[Yuan et~al.(2026{\natexlab{a}})Yuan, Liang, Chen, Wang, Li, Lin,
  Huang, Lei, Zhang, Zhang, et~al.]{yuan2026qwen}
Haoqi Yuan, Zhixuan Liang, Anzhe Chen, Ye~Wang, Haoyang Li, Pei Lin, Yiyang
  Huang, Zixing Lei, Tong Zhang, Jiazhao Zhang, et~al.
\newblock Qwen-robotmanip technical report: Alignment unlocks scale for robotic
  manipulation foundation models.
\newblock \emph{arXiv preprint}, 2026{\natexlab{a}}.

\bibitem[Yuan et~al.(2026{\natexlab{b}})Yuan, Dong, Liu, and
  Zhao]{yuan2026fastwam}
Tianyuan Yuan, Zibin Dong, Yicheng Liu, and Hang Zhao.
\newblock Fast-wam: Do world action models need test-time future imagination?
\newblock \emph{arXiv preprint}, 2026{\natexlab{b}}.

\bibitem[Zhang et~al.(2025)Zhang, Guo, Hu, Chen, Zhu, and Chen]{zhang2025upvla}
Jianke Zhang, Yanjiang Guo, Yucheng Hu, Xiaoyu Chen, Xiang Zhu, and Jianyu
  Chen.
\newblock Up-vla: A unified understanding and prediction model for embodied
  agent.
\newblock \emph{arXiv preprint}, 2025.

\bibitem[Zhang et~al.(2026{\natexlab{a}})Zhang, Luo, Hu, Chen, Guo, Liu, Xu,
  Lan, and Chen]{zhang2026uam}
Jianke Zhang, Yuanfei Luo, Yucheng Hu, Xiaoyu Chen, Yanjiang Guo, Ziyang Liu,
  Hongbin Xu, Tian Lan, and Jianyu Chen.
\newblock Uam: A dual-stream perspective on forgetting in vla training.
\newblock \emph{arXiv preprint}, 2026{\natexlab{a}}.

\bibitem[Zhang et~al.(2026{\natexlab{b}})Zhang, Hao, Tang, Zhou, Zhang, Liu,
  Li, Zhang, Zhang, Ye, et~al.]{zhang2026onevla}
Lingfeng Zhang, Xiaoshuai Hao, Yingbo Tang, Lei Zhou, Shuyi Zhang, Jinkun Liu,
  Hongsheng Li, Chenhao Zhang, Qiang Zhang, Hangjun Ye, et~al.
\newblock Onevla: A unified framework for embodied tasks.
\newblock \emph{arXiv preprint}, 2026{\natexlab{b}}.

\bibitem[Zhao et~al.(2025)Zhao, Lu, Kim, Fu, Zhang, Wu, Li, Ma, Han, Finn,
  et~al.]{zhao2025cot}
Qingqing Zhao, Yao Lu, Moo~Jin Kim, Zipeng Fu, Zhuoyang Zhang, Yecheng Wu,
  Zhaoshuo Li, Qianli Ma, Song Han, Chelsea Finn, et~al.
\newblock Cot-vla: Visual chain-of-thought reasoning for vision-language-action
  models.
\newblock In \emph{CVPR}, 2025.

\bibitem[Zhao et~al.(2023)Zhao, Kumar, Levine, and Finn]{zhao2023act}
Tony~Z. Zhao, Vikash Kumar, Sergey Levine, and Chelsea Finn.
\newblock Learning fine-grained bimanual manipulation with low-cost hardware.
\newblock \emph{arXiv preprint}, 2023.

\bibitem[Zheng et~al.(2026)Zheng, Li, Wang, Liu, Kang, Feng, Zheng, Zou, Chen,
  Zeng, et~al.]{zheng2026x}
Jinliang Zheng, Jianxiong Li, Zhihao Wang, Dongxiu Liu, Xirui Kang, Yuchun
  Feng, Yinan Zheng, Jiayin Zou, Yilun Chen, Jia Zeng, et~al.
\newblock X-vla: Soft-prompted transformer as scalable cross-embodiment
  vision-language-action model.
\newblock In \emph{ICLR}, 2026.

\bibitem[Zheng et~al.(2025)Zheng, Wang, Reed, Bjorck, Fang, Hu, Jang, Kundalia,
  Lin, Magne, et~al.]{zheng2025flare}
Ruijie Zheng, Jing Wang, Scott Reed, Johan Bjorck, Yu~Fang, Fengyuan Hu, Joel
  Jang, Kaushil Kundalia, Zongyu Lin, Loic Magne, et~al.
\newblock Flare: Robot learning with implicit world modeling.
\newblock \emph{arXiv preprint}, 2025.

\bibitem[Zhou et~al.(2026)Zhou, Zhang, Xu, Fan, Zhao, Wang, Luo, Yang, Zhu,
  Shen, et~al.]{zhou2026zerowam}
Jiaming Zhou, Qihang Zhang, Gangwei Xu, Cunxin Fan, Yujie Zhao, Ruilin Wang,
  Yiming Luo, Shuai Yang, Xing Zhu, Yujun Shen, et~al.
\newblock Zero-wam: In-context world-action modeling from human videos for
  open-ended task generalization.
\newblock \emph{arXiv preprint}, 2026.

\bibitem[Zhu et~al.(2025)Zhu, Yu, Feng, Burchfiel, Shah, and Gupta]{zhu2025uwm}
Chuning Zhu, Raymond Yu, Siyuan Feng, Benjamin Burchfiel, Paarth Shah, and
  Abhishek Gupta.
\newblock Unified world models: Coupling video and action diffusion for
  pretraining on large robotic datasets.
\newblock \emph{arXiv preprint}, 2025.

\end{thebibliography}

\clearpage
\appendix
\startcontents[appendix]
\section*{Supplementary material}
\subsection*{Contents}
\printcontents[appendix]{}{1}{\setcounter{tocdepth}{2}}
\clearpage

\section{Implementation details}
\label{sec:supp_implementation}

Both instantiations connect action learning and future prediction through layer-wise attention between experts.
Each expert maintains its own feature transformations, while attention allows it to read the permitted representations from other experts.
The future-prediction branch therefore reads action features throughout the network, rather than the final numerical action output.
Below, we describe the visual representations, attention permissions, and computation retained at deployment.

\subsection{\fdmvlm{}: architecture and visual targets}
\label{sec:supp_vlm_implementation}

The VLM-based instantiation augments the context and action experts with a generation expert of the same architecture as the action expert.
The generation expert is randomly initialized and has its own attention projections, feed-forward layers, and normalization parameters.
For each camera view $k$, let $O_t^{(k)}$ denote its image within the current observation $O_t$.
The generation expert receives visual context from $O_{t-15}^{(k)}$ and $O_t^{(k)}$, and predicts $O_{t+15}^{(k)}$, where offsets are measured in recorded frames.
The context expert processes only the current images and task instruction.
Thus, the additional temporal context supports future prediction during training without becoming an input requirement for the deployed policy.

A frozen Wan VAE encodes each image independently into a 48-channel latent grid with a spatial downsampling factor of 16.
Independent encoding keeps the future target separate from the historical and current inputs.
For the generation expert, a learned channel projection followed by a strided convolution converts the two context grids into a shorter sequence of visual tokens.
After transformer processing, features at corresponding spatial locations are averaged across the two temporal positions separately for each view.
A transposed convolution restores the latent-grid resolution, and normalization followed by a linear projection produces the predicted future latent.
The visual objective averages latent reconstruction errors over valid camera views.
This is direct regression: the generation expert predicts a clean latent in one forward pass, with a fixed zero time condition and no visual denoising loop.

\subsection{\fdmvgm{}: architecture and visual targets}
\label{sec:supp_vgm_implementation}

The VGM-based instantiation pairs a generation expert with an action expert and connects corresponding transformer layers through attention.
Their attention heads have matching dimensions, allowing action and video features to interact while retaining expert-specific transformations.
Camera views are composed spatially before VAE encoding, so one video stream represents the multi-view scene.
The generation expert processes both the current observation and future video; attention permissions distinguish their roles in action prediction.

During training, the current-frame latent is kept clean, while future-video latents and the demonstrated action chunk are corrupted using independently sampled noise and flow-matching times.
The future-video tokens read the hidden representations of the noised action chunk at that training time.
Video supervision is applied only to future latents, excluding the fixed current frame and masking padded temporal targets.
The joint objective in \Cref{eq:total} uses $\lambda_a=\lambda_v=1$ in both instantiations.

\subsection{Attention permissions and backward supervision}
\label{sec:supp_attention}

\Cref{fig:supp_attention_mask} illustrates the asymmetric attention mask, and \Cref{tab:supp_attention} details the token groups in each instantiation.
In the VLM-based model, context features are available to both downstream experts, and generation tokens additionally read action features.
In the VGM-based model, current-observation tokens attend only to the current observation within the visual--action sequence; action tokens read the current observation and other action tokens, while future-video tokens read all three groups.
Future-video attention is bidirectional across future positions.
Language conditioning is supplied separately to the VGM experts.
These permissions hold at every layer, keeping both action features and the observation features they read independent of generation-branch tokens.

\begin{figure}[!htbp]
    \centering
    \begin{minipage}[c]{0.49\linewidth}
        \centering
        \includegraphics[width=\linewidth]{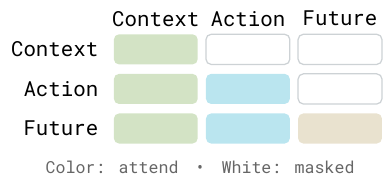}
    \end{minipage}\hfill
    \begin{minipage}[c]{0.47\linewidth}
    \caption{\textbf{Asymmetric attention mask in \model{}.}
    Rows denote queries; columns denote keys/values.
    Colored blocks allow attention; white blocks mask it.
    Context/Future denote context/generation tokens in \fdmvlm{} and current-observation/future-video tokens in \fdmvgm{}.}
    \label{fig:supp_attention_mask}
    \end{minipage}
\end{figure}

\begin{table}[htbp]
    \centering
    \caption{\textbf{Attention permissions between token groups.}
    Rows denote queries and columns denote keys/values; a checkmark permits attention.
    VLM generation tokens originate from historical and current image latents and are used to predict the future target.
    VGM language conditioning is separate from the groups shown here.}
    \label{tab:supp_attention}
    \small
    \begin{tabular}{llccc}
        \toprule
        Model & Query group & \multicolumn{3}{c}{Key/value group} \\
        \cmidrule(lr){3-5}
        & & Context & Action & Generation \\
        \midrule
        \multirow{3}{*}{VLM} & Context & \cmark & \xmark & \xmark \\
        & Action & \cmark & \cmark & \xmark \\
        & Generation & \cmark & \cmark & \cmark \\
        \midrule
        & & Current observation & Action & Future video \\
        \midrule
        \multirow{3}{*}{VGM} & Current observation & \cmark & \xmark & \xmark \\
        & Action & \cmark & \cmark & \xmark \\
        & Future video & \cmark & \cmark & \cmark \\
        \bottomrule
    \end{tabular}
\end{table}

The same connection carries supervision in the reverse direction during optimization.
The future loss differentiates through the action keys and values read by the generation expert and updates the action features that produce them.
Action imitation simultaneously supervises those features through the action velocity head.
For the stop-gradient control in \Cref{sec:ablation}, the generation-facing action features retain their forward values while their backward connection is blocked.
Gradients through shared context are a separate path, so this control specifically isolates supervision through the action interface.

\subsection{Action-only deployment and optional visual prediction}
\label{sec:supp_deployment}

At deployment, the VLM-based policy uses its context and action experts to denoise an action chunk; the additional generation expert and its VAE processing are unnecessary for control.
The VGM-based policy first processes the current observation and caches its keys and values at each transformer layer.
Each action-denoising step reuses this cache, updating only the action stream without allocating or processing future-video tokens.
The generation expert's observation processing is retained because it supplies the visual context for action generation.

Optional visual rollout uses the same action-to-future connection, with sampling schedules specific to each instantiation.
Given a supplied action chunk, its representations condition future prediction with the action flow-matching time $\tau_a$ set to zero.
For this optional rollout, the VLM-based implementation samples an action chunk before predicting future-image latents in a single pass.
The VGM-based implementation can denoise actions and future video together using their respective schedules, restoring the fixed current-frame latent after each video update.
Throughout this process, generation-branch tokens read action features while the action stream remains independent of future predictions.

\FloatBarrier

\subsection{Implementation of FastWAM-Joint}
\label{sec:supp_wam_implementation}

We implement FastWAM-Joint, the future-conditioned WAM used throughout the real-world task-success comparison (\Cref{sec:sim_results}), the future-perturbation analysis (\Cref{sec:future_error_sensitivity}) and the efficiency comparison (\Cref{sec:efficiency}), with a generation expert and an action expert connected through MoT attention.
We retain the notation $C=(O_t,L)$ for the current observation and task instruction, $A$ for the action chunk, and $V$ for the clean future-video latents.
FastWAM-Joint realizes the future-to-action dependency in \Cref{tab:para_compare}: action prediction reads the evolving future-video representations during joint denoising.
The following description concerns this implementation of the WAM paradigm.

\paragraph{Visual representations and action features.}
The generation expert uses the Wan video backbone, and a frozen Wan VAE encodes the spatially composed camera views into video latents.
The current-frame latent provides the visual condition, while the subsequent latents represent $V$.
Spatial patches of these latents form the generation expert's tokens.
The action expert projects each vector in the noised action chunk $A_{\tau_a}$ into an action token and uses the action flow-matching time $\tau_a$ for time conditioning.
Both experts receive text features of the task instruction $L$ through their own cross-attention layers.
Their transformer layers have matching attention-head dimensions and are connected at corresponding depths, while retaining separate projections, normalization layers, and feed-forward transformations.
An action velocity head produces $\widehat u_a$ from the action features, and the generation expert predicts the future-video velocity.

\paragraph{Future-to-action attention.}
At every layer, action tokens attend to the current-observation tokens, future-video tokens, and other action tokens.
The generation expert processes the visual sequence independently of action tokens.
Within the visual sequence, current-observation tokens attend only to the current observation, while future-video tokens attend to the current observation and all future positions.
With rows denoting queries and columns denoting keys/values in the order (current observation, action, future video), these permissions are
\begin{equation}
\label{eq:wam_attention}
M_{\text{FastWAM-Joint}}=
\begin{pmatrix}
1 & 0 & 0 \\
1 & 1 & 1 \\
1 & 0 & 1
\end{pmatrix},
\end{equation}
where each entry denotes an entire allowed or blocked attention block.
Thus, action features incorporate the current denoising estimate of the future through the generation expert's hidden states.
Compared with the attention permissions of \fdmvgm{} in \Cref{tab:supp_attention}, the cross-expert connection is reversed: FastWAM-Joint actions read future-video representations, whereas FDM future-video tokens read action features.

\paragraph{Joint training.}
Training corrupts $V$ according to \Cref{eq:fm} and independently corrupts the demonstrated action chunk as
\begin{equation}
\label{eq:wam_action_noise}
A_{\tau_a}=(1-\tau_a)A+\tau_a\epsilon_a,
\qquad \epsilon_a\sim\mathcal{N}(0,I).
\end{equation}
The action and video flow-matching times, $\tau_a$ and $\tau_v$, are sampled independently during training.
The current-frame latent remains clean and receives a zero time condition.
A joint forward pass predicts the action velocity $\widehat u_a$ and the future-video velocity, whose flow-matching targets are $\epsilon_a-A$ and $\epsilon_v-V$, respectively.
The future-prediction loss $\mathcal{L}_{\mathrm{world}}=\mathcal{L}_{\mathrm{vid}}$ excludes the current frame and averages errors over valid future targets.
The action loss $\mathcal{L}_{\mathrm{act}}$ similarly masks invalid action entries.
For velocity supervision, it compares $\widehat u_a$ with $\epsilon_a-A$; for clean-action supervision, it compares $A_{\tau_a}-\tau_a\widehat u_a$ with $A$.
Both forms train the same action velocity head used during sampling.
The total objective combines the two losses with $\lambda_a$ and $\lambda_v$ as in \Cref{eq:total}, with any time-dependent weighting applied within the respective loss.
Through the permitted attention connection, action-loss gradients can update the visual representations read by the action expert.

\paragraph{Joint denoising at inference.}
Inference starts from Gaussian noise for both the action chunk and future-video latents, while fixing the current-frame latent to the encoded observation.
The two streams use the same descending flow-matching times, $1=\tau^{(0)}>\cdots>\tau^{(K)}=0$, where $K$ is the number of denoising steps.
At step $j$, a joint forward pass evaluates both experts with $\tau_a=\tau_v=\tau^{(j)}$.
Writing their velocity predictions as $\widehat u_a^{(j)}$ and $\widehat u_v^{(j)}$, the updates are
\begin{equation}
\label{eq:wam_joint_sampling}
\begin{aligned}
A_{\tau^{(j+1)}} &= A_{\tau^{(j)}}
+\bigl(\tau^{(j+1)}-\tau^{(j)}\bigr)\widehat u_a^{(j)}, \\
V_{\tau^{(j+1)}} &= V_{\tau^{(j)}}
+\bigl(\tau^{(j+1)}-\tau^{(j)}\bigr)\widehat u_v^{(j)}.
\end{aligned}
\end{equation}
The current-frame latent is restored after every video update.
Consequently, future conditioning operates throughout sampling: the action expert reads partially denoised future representations at each step, without waiting for a completed video rollout.
Returning only the action chunk omits video decoding, but still requires future-video latent computation during denoising.
This dependence explains both why perturbations to future latents can change FastWAM-Joint actions in \Cref{sec:future_error_sensitivity} and why future generation contributes to its action inference cost in \Cref{sec:efficiency}.

\FloatBarrier

\section{Simulation training and evaluation configurations}
\label{sec:supp_sim_config}

We report the simulation training configurations for \fdmvlm{} and \fdmvgm{} on LIBERO and RoboTwin 2.0.
Training settings and dataset splits are taken from the saved configurations and training logs.
We use the following evaluation protocol for all reported experiments.

\subsection{Training configurations}
\label{sec:supp_sim_training}

All four training runs use seed 42 and eight GPU processes, with 16 samples per device per microbatch.
The effective global batch size additionally accounts for gradient accumulation, as listed in \Cref{tab:sim_training}.
We use AdamW with $(\beta_1,\beta_2)=(0.9,0.95)$ and $\epsilon=10^{-8}$, gradient clipping at a norm of 1.0, BF16 precision, and DeepSpeed ZeRO-2.
Gradient checkpointing is disabled in all runs.
The learning rate follows cosine decay after warmup.
Offline validation data are held out at the episode level using split seed 42; the episode counts in \Cref{tab:sim_training} follow the actual training logs.
Offline validation measures held-out training objectives and is separate from closed-loop success-rate evaluation.

\begin{table}[htbp]
    \caption{Simulation training configurations. Counts and final optimizer steps follow the training logs. Validation intervals refer to offline validation.}
    \label{tab:sim_training}
    \centering
    \small
    \setlength{\tabcolsep}{4pt}
    \renewcommand{\arraystretch}{1.1}
    \begin{tabular}{@{}lcccc@{}}
        \toprule
        & \multicolumn{2}{c}{\fdmvlm{}} & \multicolumn{2}{c}{\fdmvgm{}} \\
        \cmidrule(lr){2-3} \cmidrule(lr){4-5}
        Setting & LIBERO & RoboTwin & LIBERO & RoboTwin \\
        \midrule
        Training episodes & 1,659 & 26,950 & 1,659 & 27,225 \\
        Validation episodes & 34 & 550 & 34 & 275 \\
        Validation fraction & 2\% & 2\% & 2\% & 1\% \\
        Training steps & 30,000 & 60,000 & 22,000 & 29,375 \\
        Warmup steps & 1,000 & 1,000 & 1,100 & 1,468 \\
        Validation interval (steps) & 2,000 & 2,000 & 200 & 500 \\
        Checkpoint interval (steps) & 2,000 & 2,000 & 10,000 & 2,500 \\
        \midrule
        Per-device batch size & 16 & 16 & 16 & 16 \\
        Gradient accumulation & 1 & 2 & 1 & 8 \\
        Effective global batch size & 128 & 256 & 128 & 1,024 \\
        Peak learning rate & $2.5\times10^{-5}$ & $2.5\times10^{-5}$ & $10^{-4}$ & $10^{-4}$ \\
        Minimum learning rate & $0$ & $0$ & $10^{-6}$ & $10^{-6}$ \\
        Weight decay & $10^{-10}$ & $10^{-10}$ & $10^{-2}$ & $10^{-2}$ \\
        \midrule
        Camera views & 2 & 3 & 2 & 3 \\
        Proprioceptive state input & False & False & False & False \\
        Environment action dimension & 7 & 14 & 7 & 14 \\
        Action prediction horizon & 15 & 50 & 32 & 32 \\
        Normalization & Min--Max & Min--Max & Min--Max & Z-score \\
        \bottomrule
    \end{tabular}
\end{table}

\FloatBarrier

\paragraph{\fdmvlm{}.}
The saved configurations initialize the policy from \texttt{pi05\_base\_pytorch}, with a Gemma-2B language backbone, a Gemma-300M action expert, and a So400m/14 vision encoder.
The additional generation expert is randomly initialized and has not undergone pretraining.
The maximum text length is 200 tokens, and action vectors are internally padded to 32 dimensions.
LIBERO uses external and wrist camera views, while RoboTwin uses head, left-wrist, and right-wrist views, all at $224\times224$ resolution.
Proprioceptive state input is disabled on both LIBERO and RoboTwin.
Both datasets use Min--Max normalization.
Compilation is enabled for these runs.
The generation-expert configuration specifies the Wan VAE from Wan2.2, a resolution value of 256, a scale factor of four, and a loss weight of 1.0.

\paragraph{\fdmvgm{}.}
The video backbone is initialized from Wan2.2-TI2V-5B.
The action expert is randomly initialized and has not undergone pretraining; the initialization checkpoint loaded by the saved configurations stores these random initial weights.
Both experts have 30 layers, with hidden dimensions of 3072 for the video backbone and 1024 for the action expert.
Training uses cached UMT5 text features with a maximum text length of 128 tokens, and compilation is disabled.
Action and future-prediction losses both have unit weight.
Future frames are sampled every four steps, and the action prediction horizon is 32 for both benchmarks.
The video and action schedulers each use 1,000 training timesteps and a shift of 5.0 for training and inference; action timesteps use the configured \texttt{beta\_tau} sampling scheme.
Proprioceptive state input is disabled on both benchmarks.
LIBERO uses Min--Max normalization, while RoboTwin uses Z-score normalization.

\paragraph{Multi-view video inputs.}
For LIBERO, the VGM configuration uses two camera views, center-cropped and resized to $224\times224$ per view, then concatenated horizontally.
For RoboTwin, the configured per-view size is $240\times320$.
The current RoboTwin layout implementation assembles a $384\times320$ canvas, with a $256\times320$ head-camera view above two $128\times160$ wrist-camera views.
These canvas dimensions are distinct from the configured per-view preprocessing size.

\subsection{Default inference and evaluation protocol}
\label{sec:supp_sim_evaluation}

The current evaluation defaults use the same environment protocol and action execution horizon for both model families within each benchmark (\Cref{tab:sim_evaluation}).
The action prediction horizon and normalization remain specific to the model checkpoint.
In particular, RoboTwin uses Min--Max normalization for \fdmvlm{} and Z-score normalization for \fdmvgm{}; evaluation-template fields do not override the normalization loaded by the policy server.
The default sampling procedure uses 10 action denoising steps and evaluation seed 42.

\begin{table}[htbp]
    \caption{Current default simulation evaluation settings, shared by both model families except for the prediction horizon. Historical evaluation runs require verification against these defaults.}
    \label{tab:sim_evaluation}
    \centering
    \small
    \setlength{\tabcolsep}{5pt}
    \renewcommand{\arraystretch}{1.1}
    \begin{tabular}{@{}lcc@{}}
        \toprule
        Setting & LIBERO & RoboTwin \\
        \midrule
        Number of tasks & 40 & 50 \\
        Episodes per task per condition & 50 & 100 \\
        Evaluation seed & 42 & 42 \\
        Action denoising steps & 10 & 10 \\
        \fdmvlm{} action prediction horizon & 15 & 50 \\
        \fdmvgm{} action prediction horizon & 32 & 32 \\
        Action execution horizon & 10 & 24 \\
        \bottomrule
    \end{tabular}
\end{table}

\FloatBarrier

\paragraph{LIBERO.}
The default evaluation covers Spatial, Object, Goal, and LIBERO-Long (LIBERO-10), with 50 episodes for each of the 40 tasks, totaling 2,000 episodes per complete evaluation.
Each episode starts from an official initial state and includes 30 initial settling steps.
The control-step limits are 400 for Spatial, Object, and Goal, and 700 for Long, in addition to the settling steps.
The policy receives the task description as its task instruction and outputs end-effector position and rotation increments with a binarized gripper command.
Real-time chunking is disabled by default.

\paragraph{RoboTwin.}
The default evaluation covers all 50 tasks, with 100 episodes per task under each of the clean and randomized conditions, totaling 5,000 episodes per condition.
Task environments determine initialization and episode limits.
The policy uses absolute joint-position commands (\texttt{action\_type=qpos}) and the unseen-instruction setting (\texttt{instruction\_type=unseen}).
By default, repeated observation acquisition is skipped while executing an action chunk.
Per-task success rates are aggregated across the 50 tasks; the clean/randomized averaging convention is specified in \Cref{sec:supp_robotwin_results}.

\section{Real-world training hyperparameters}
\label{sec:supp_hyperparameters}

We summarize the training configurations used for \fdmvlm{} and \fdmvgm{} in the real-world experiments.
Both instantiations use the Wan VAE associated with the Wan2.2-5B backbone to encode visual inputs and prediction targets and to decode predicted latents.
Both configurations use AdamW with $(\beta_1,\beta_2)=(0.9,0.95)$ and $\epsilon=10^{-8}$, a gradient clipping norm of $1.0$, and cosine learning-rate decay after 1,000 warmup steps.
Training uses eight NVIDIA H100 GPUs with a per-device batch size of eight, giving an effective batch size of 64.
Training uses DeepSpeed ZeRO-2, with gradient checkpointing disabled.
The training budget is 30,000 optimizer steps.
The action execution horizon is 25: both policies execute 25 actions from each predicted action chunk before replanning.

\subsection{\fdmvlm{}}
\label{sec:supp_vla_hyperparameters}

For the $\pi_{0.5}$-based policy, we use a learning rate of $2.5\times10^{-5}$, decayed to zero, and weight decay of $10^{-10}$.
The policy takes three camera images at $224\times224$ resolution, with proprioceptive state input disabled.
The action prediction horizon is 50 steps, and the model action dimension is 20.
The action and future-prediction losses both have unit weight.
Action denoising uses 10 steps.
\Cref{tab:vla_hyperparameters} lists the VLM-based settings.

\begin{table}[htbp]
    \caption{Training hyperparameters for \fdmvlm{} in the real-world experiments.}
    \label{tab:vla_hyperparameters}
    \begin{center}
    \small
    \begin{tabular}{ll}
        \toprule
        Hyperparameter & Value \\
        \midrule
        Peak learning rate & $2.5\times10^{-5}$ \\
        Minimum learning rate & $0$ \\
        Weight decay & $10^{-10}$ \\
        Training steps / warmup steps & 30,000 / 1,000 \\
        Per-device / effective batch size & 8 / 64 \\
        Camera views & 3 \\
        Per-view image resolution & $224\times224$ \\
        Proprioceptive state input & False \\
        Action prediction horizon & 50 \\
        Action execution horizon & 25 \\
        Model action dimension & 20 \\
        Action / future-prediction loss weight & 1.0 / 1.0 \\
        Action denoising steps & 10 \\
        \bottomrule
    \end{tabular}
    \end{center}
\end{table}

\FloatBarrier

\subsection{\fdmvgm{}}
\label{sec:supp_wam_hyperparameters}

For the Wan-based policy, we use a learning rate of $10^{-4}$, decayed to $10^{-6}$, and weight decay of $10^{-2}$.
Each camera view is resized to $240\times320$ before multi-camera composition, and proprioceptive state input is disabled.
The action prediction horizon is 32 steps, and future training frames are sampled with a stride of four.
The model action dimension is 20.
The action and future-prediction losses both have unit weight.
Training uses cached text embeddings, and action denoising uses 10 steps.
\Cref{tab:wam_hyperparameters} lists the VGM-based settings.

\begin{table}[htbp]
    \caption{Training hyperparameters for \fdmvgm{} in the real-world experiments.}
    \label{tab:wam_hyperparameters}
    \begin{center}
    \small
    \begin{tabular}{ll}
        \toprule
        Hyperparameter & Value \\
        \midrule
        Peak learning rate & $10^{-4}$ \\
        Minimum learning rate & $10^{-6}$ \\
        Weight decay & $10^{-2}$ \\
        Training steps / warmup steps & 30,000 / 1,000 \\
        Per-device / effective batch size & 8 / 64 \\
        Gradient accumulation steps & 1 \\
        Per-view image resolution & $240\times320$ \\
        Proprioceptive state input & False \\
        Action prediction horizon & 32 \\
        Action execution horizon & 25 \\
        Model action dimension & 20 \\
        Future-frame stride & 4 \\
        Action / future-prediction loss weight & 1.0 / 1.0 \\
        Action denoising steps & 10 \\
        \bottomrule
    \end{tabular}
    \end{center}
\end{table}

\FloatBarrier

\subsection{Action-to-future ablation protocol}
\label{sec:supp_ablation_protocol}

The ablation in \Cref{sec:ablation} compares A. Action-only, B. Context-only future, C. Stop-gradient action conditioning, and D. Full \model{} within each backbone family.

\paragraph{Matched training settings.}
Within each backbone and training seed, the four variants share the policy initialization, training data, action objective and weight, number of optimizer steps, effective batch size, and optimizer.
B, C, and D additionally share the generation-expert architecture and initialization, future targets, and future-loss weight.

\paragraph{Stop-gradient implementation.}
In C, the stop-gradient operation applies to every action-feature interface read by the future-prediction branch, while leaving the action velocity head's access to $H_a$ intact.
All other attention permissions and shared-parameter update paths are held fixed between C and D.
Gradients through shared context processing may remain in C; the comparison therefore isolates future-loss gradients through $H_a$.
For \fdmvgm{}, all variants retain the current-observation computation required by the action expert.

\paragraph{Evaluation.}
All four variants use the same evaluation protocol and action-only inference on Drawer, Box, Shuttlecock, Cup, and Block (\Cref{sec:supp_real_tasks}).
\Cref{tab:ablation} reports per-task closed-loop success rates and their unweighted average separately for the VLM- and VGM-based instantiations, with comparisons using matched training seeds.

\section{Real-world setup and task descriptions}
\label{sec:supp_real_tasks}

We evaluate five manipulation tasks covering sequential object handling, stacking, cloth manipulation, and insertion.
The descriptions below specify the task goals and constituent operations for the evaluation in \Cref{sec:real_robot}.

\paragraph{Hardware and teleoperation.}
We use a dual-arm ARX-X5 robot and teleoperate it using a PICO 4 Ultra VR headset.
The camera setup consists of one RealSense D435 providing a third-person view of the workspace and two RealSense D405 cameras, one mounted on each arm, providing left- and right-wrist views.
\Cref{fig:expset} illustrates the hardware setup.

\begin{figure}[htbp]
    \centering
    \includegraphics[width=0.58\linewidth]{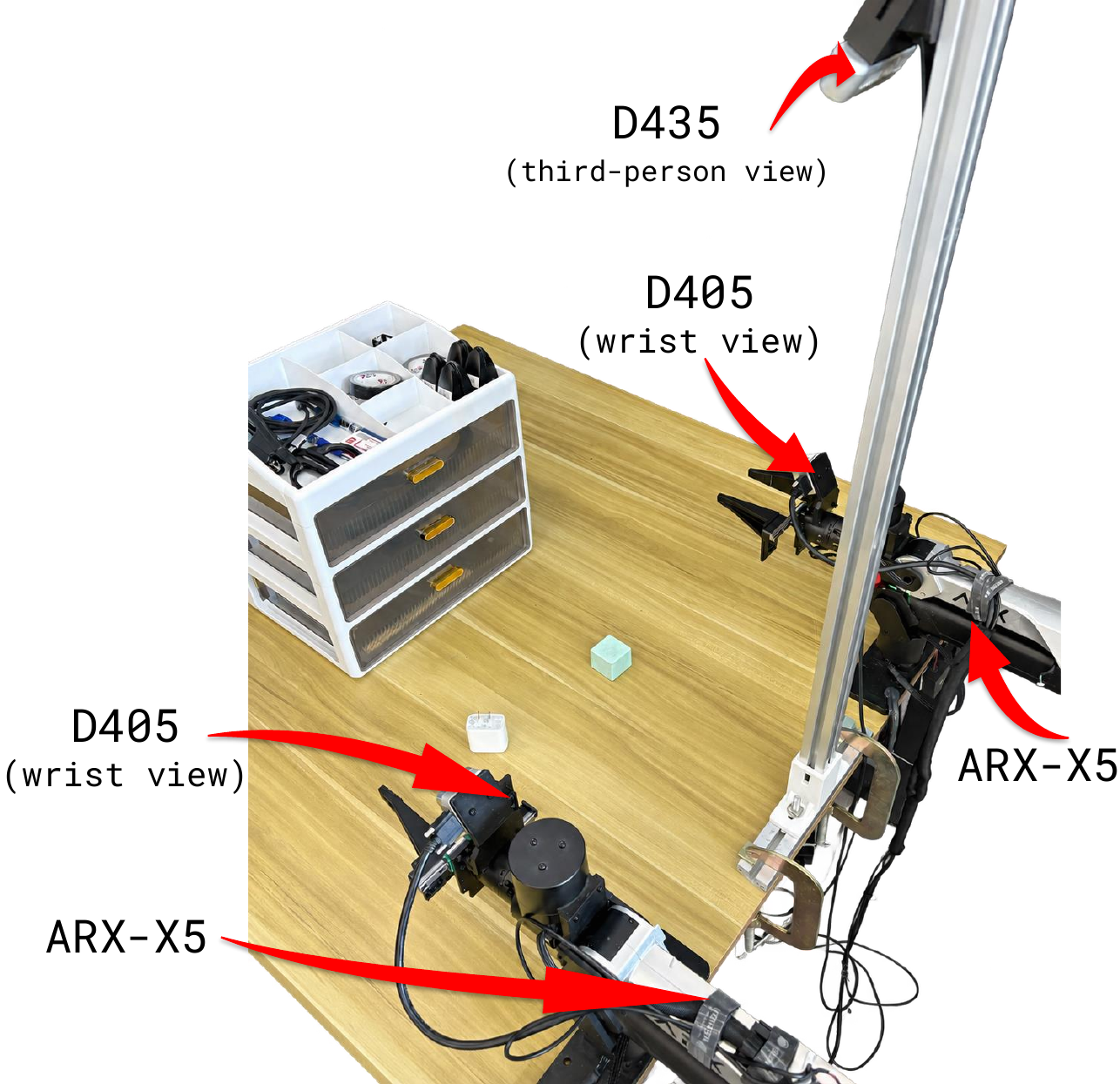}
    \caption{\textbf{Real-world experimental setup.}
    Two ARX-X5 robot arms are equipped with wrist-mounted D405 cameras, while a D435 camera provides a third-person view of the workspace.}
    \label{fig:expset}
\end{figure}

\paragraph{Data collection and action parameterization.}
We collect 150 demonstration trajectories per task using VR teleoperation, reserving three trajectories for offline validation and using the remaining 147 for training.
Incremental position and rotation commands from the VR interface are converted into robot joint commands through inverse kinematics.
The control frequency is 30\,Hz.
Each arm uses a 10-dimensional action vector comprising three position components, a six-dimensional rotation representation, and one gripper command, yielding a 20-dimensional bimanual action space.
Actions are normalized using task-specific mean and standard deviation statistics.

\paragraph{Evaluation protocol.}
Real-world evaluation follows a blinded protocol: human evaluators are unaware of the identity of the model being evaluated.
Task success is determined by a human evaluator based on whether the specified task goal is achieved.
A trial is counted as a failure if it times out or a human evaluator judges that the robot cannot continue to complete the task.
Object placements vary across demonstrations and evaluation trials, with slightly different placement distributions during training and evaluation.
The task-specific placement variations and goals are described below.

\Needspace{0.34\textheight}
\subsection{Drawer packing}
\label{sec:supp_drawer_packing}

The goal is to place both a charger and a block inside a drawer and leave the drawer closed (\Cref{fig:real_drawer_packing}).
The drawer is initially closed, and both objects are outside the drawer in the workspace.
The charger and block positions are randomized, while the drawer base position varies slightly.
The robot first grasps the drawer handle and pulls outward to expose the interior.
It then picks up each object, moves it over the open drawer, and releases it inside.
Finally, the robot pushes the drawer back into the cabinet.
This sequence couples articulated-object manipulation with pick-and-place: the opening must provide sufficient clearance for insertion, and the placed objects must not obstruct the drawer's closing motion.
The desired final state has both objects contained in the drawer and the drawer fully closed.

\begin{figure}[H]
    \centering
    \includegraphics[width=0.16\linewidth]{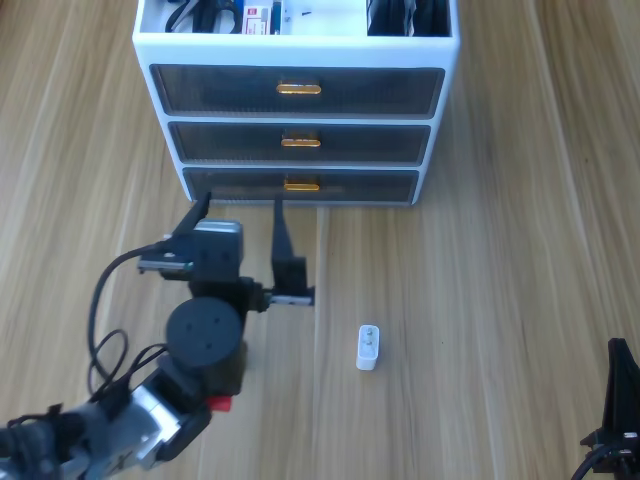}\hfill%
    \includegraphics[width=0.16\linewidth]{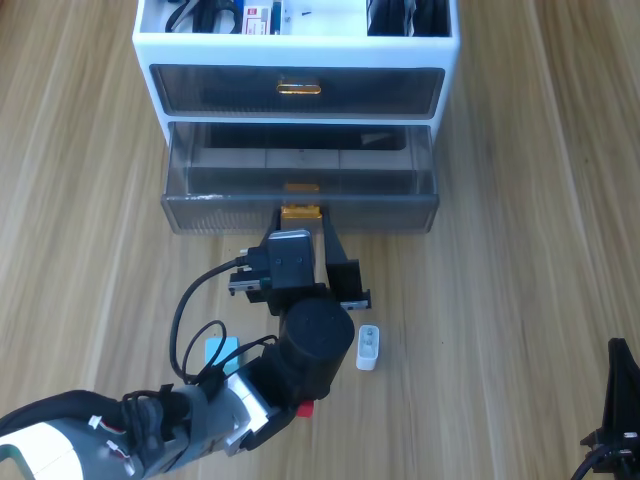}\hfill%
    \includegraphics[width=0.16\linewidth]{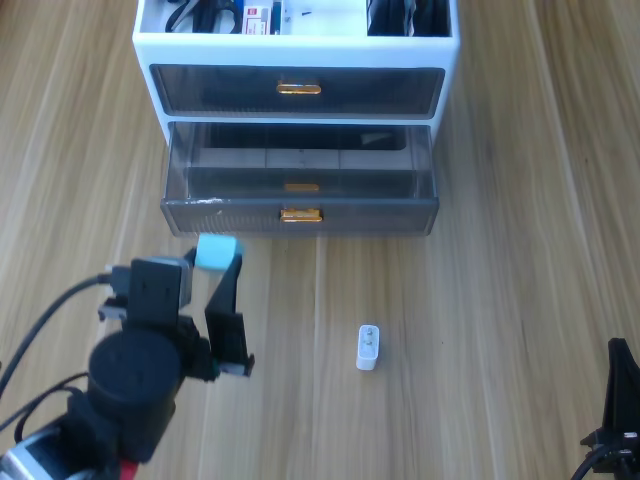}\hfill%
    \includegraphics[width=0.16\linewidth]{figure/real-task/box/frame_000911.png}\hfill%
    \includegraphics[width=0.16\linewidth]{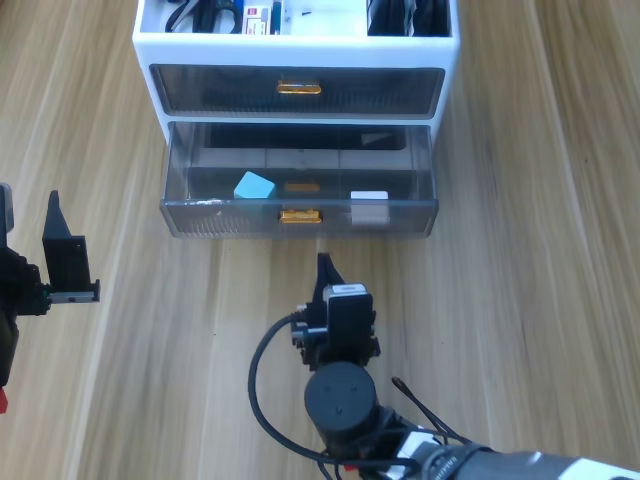}\hfill%
    \includegraphics[width=0.16\linewidth]{figure/real-task/box/frame_001176.png}%
    \caption{\textbf{Drawer packing.} The task requires opening the drawer, placing both the charger and block inside, and closing the drawer. The six frames are ordered chronologically from left to right.}
    \label{fig:real_drawer_packing}
\end{figure}

\Needspace{0.34\textheight}
\subsection{Box packing and closing}
\label{sec:supp_box_packing}

The goal is to pack a scarf and a single flower into a box and close its lid (\Cref{fig:real_box_packing}).
Initially, the box is open, with the scarf and flower lying on the tabletop beside it.
The positions of the box, flower, and scarf vary slightly.
The robot first lifts the scarf into the box and releases it within the box interior.
It then grasps the flower by its stem, transfers it over the box, and places it on top of the scarf.
After both objects have been deposited, the robot brings the hinged lid down over the opening.
The task combines handling a deformable textile with positioning an elongated object and manipulating a lid.
The contents must fit within the box so that neither the scarf nor the flower obstructs closure; the desired final state has both objects enclosed and the lid closed.

\begin{figure}[H]
    \centering
    \includegraphics[width=0.16\linewidth]{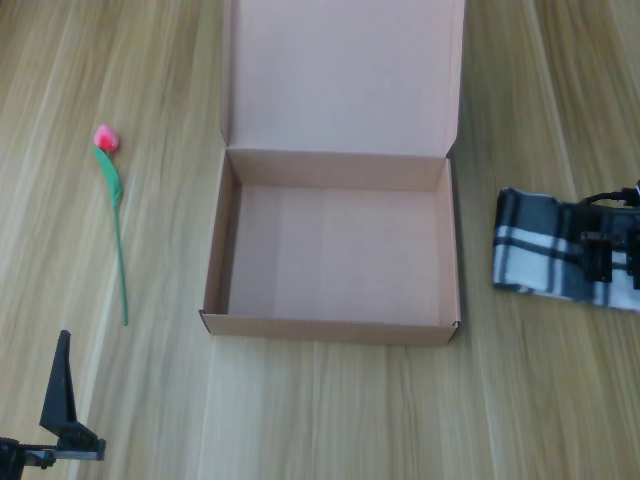}\hfill%
    \includegraphics[width=0.16\linewidth]{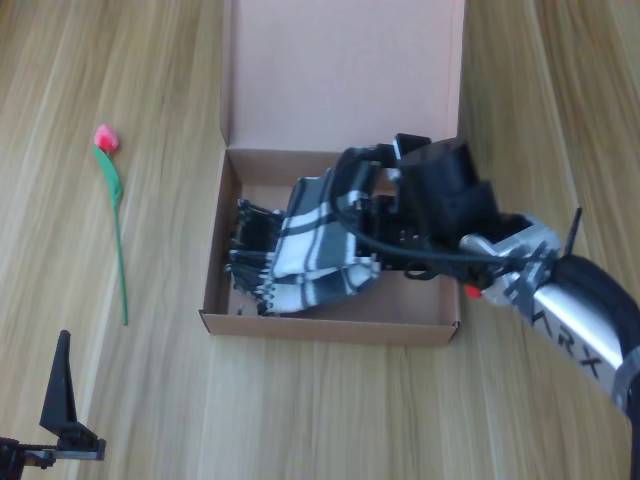}\hfill%
    \includegraphics[width=0.16\linewidth]{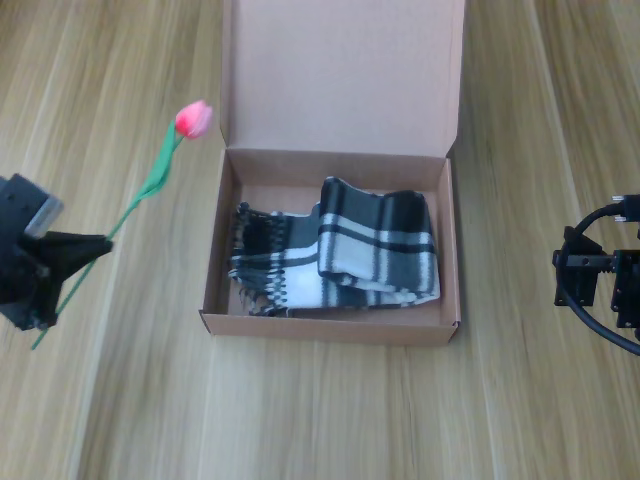}\hfill%
    \includegraphics[width=0.16\linewidth]{figure/real-task/gift/frame_000530.png}\hfill%
    \includegraphics[width=0.16\linewidth]{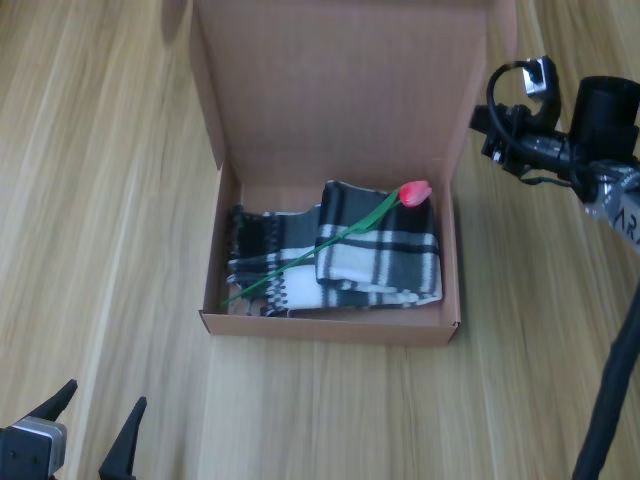}\hfill%
    \includegraphics[width=0.16\linewidth]{figure/real-task/gift/frame_001082.png}%
    \caption{\textbf{Box packing and closing.} The robot places the scarf in the box, adds the flower, and closes the lid. The six frames are ordered chronologically from left to right.}
    \label{fig:real_box_packing}
\end{figure}

\Needspace{0.34\textheight}
\subsection{Shuttlecock insertion}
\label{sec:supp_shuttlecock_insertion}

The goal is to insert two shuttlecocks into a shuttlecock tube (\Cref{fig:real_shuttlecock_insertion}).
The tube and the two shuttlecocks are initially separate on the tabletop.
The tube position varies slightly, and the shuttlecocks are placed in front of the right arm with varying positions and orientations.
One arm grasps and holds the tube horizontally, presenting its open end to the other arm.
The other arm picks up a shuttlecock, aligns its narrow end with the opening, and advances it into the tube.
The robot then repeats the pickup, alignment, and insertion sequence for the second shuttlecock while maintaining the tube's pose.
This task requires coordination between the arm holding the receptacle and the arm performing insertion, as well as alignment of each shuttlecock with the tube axis.
The desired final state has both shuttlecocks inside the tube.

\begin{figure}[H]
    \centering
    \includegraphics[width=0.16\linewidth]{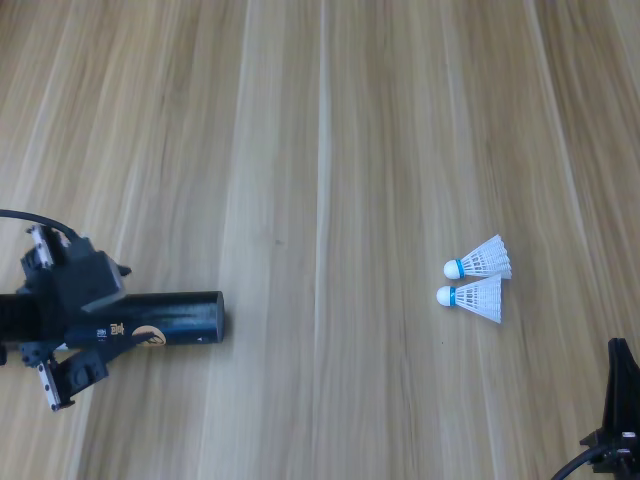}\hfill%
    \includegraphics[width=0.16\linewidth]{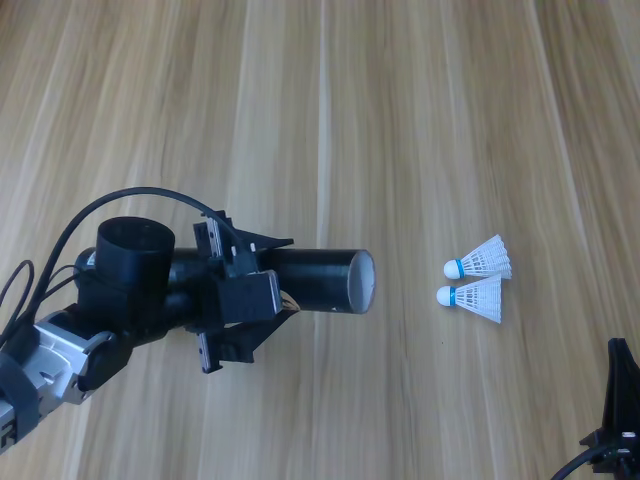}\hfill%
    \includegraphics[width=0.16\linewidth]{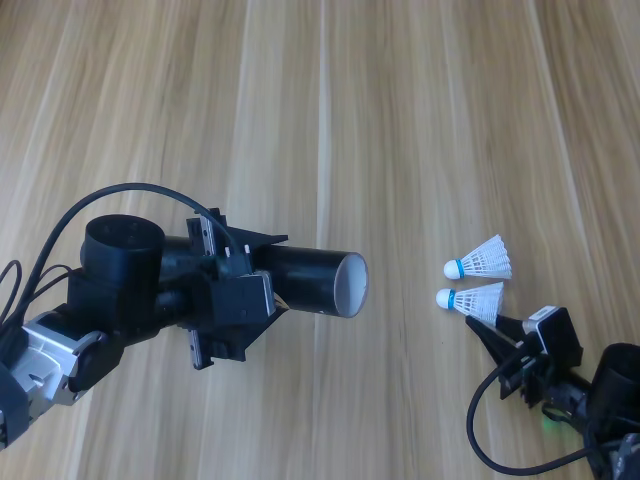}\hfill%
    \includegraphics[width=0.16\linewidth]{figure/real-task/badminton/frame_000230.png}\hfill%
    \includegraphics[width=0.16\linewidth]{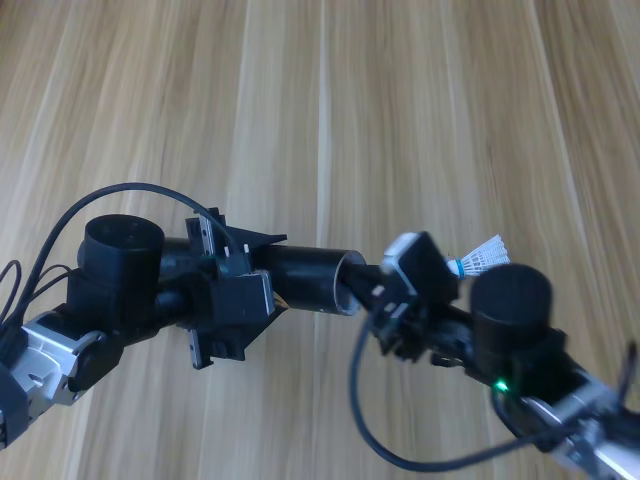}\hfill%
    \includegraphics[width=0.16\linewidth]{figure/real-task/badminton/frame_000439.png}%
    \caption{\textbf{Shuttlecock insertion.} One arm holds the tube while the other inserts the two shuttlecocks sequentially. The six frames are ordered chronologically from left to right.}
    \label{fig:real_shuttlecock_insertion}
\end{figure}

\Needspace{0.34\textheight}
\subsection{Paper-cup stacking}
\label{sec:supp_cup_stacking}

The goal is to build a two-tier structure with three paper cups in the bottom row and two cups in the upper row (\Cref{fig:real_cup_stacking}).
Initially, two cups are placed on the left and three on the right, with randomized positions within their respective sides.
Starting from cups distributed across the tabletop, the robot picks up and arranges three cups side by side to form the base.
It then lifts each of the remaining two cups and places it above a gap between adjacent base cups.
Each upper cup must be supported by the cups beneath it, so the spacing of the base determines the available placement locations.
The robot must control both the approach and release to avoid displacing the lightweight cups or disturbing an earlier placement.
The desired final state is a freestanding five-cup structure that remains stable after the grippers withdraw.

\begin{figure}[H]
    \centering
    \includegraphics[width=0.16\linewidth]{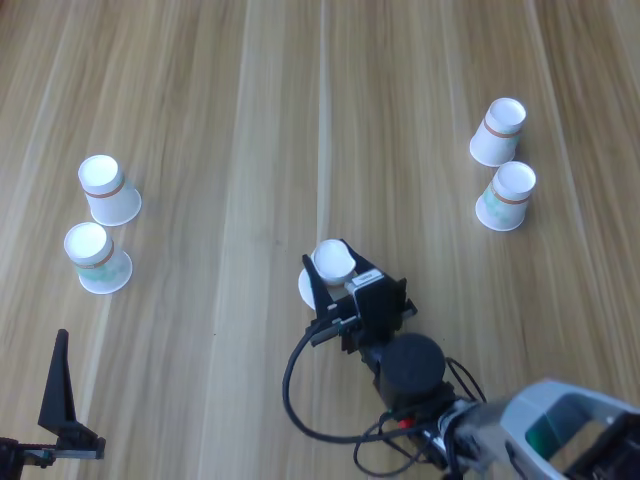}\hfill%
    \includegraphics[width=0.16\linewidth]{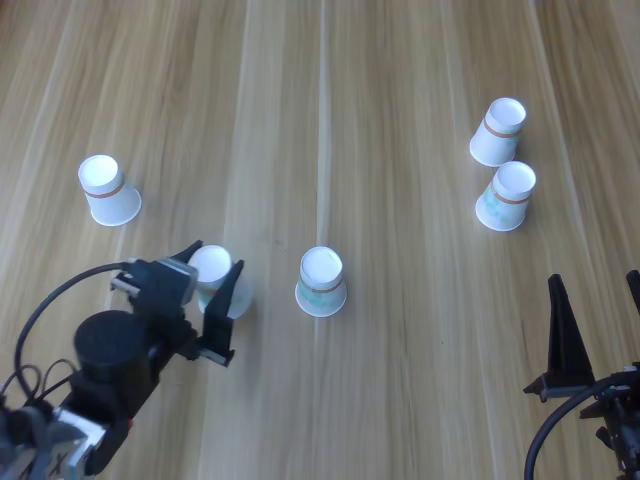}\hfill%
    \includegraphics[width=0.16\linewidth]{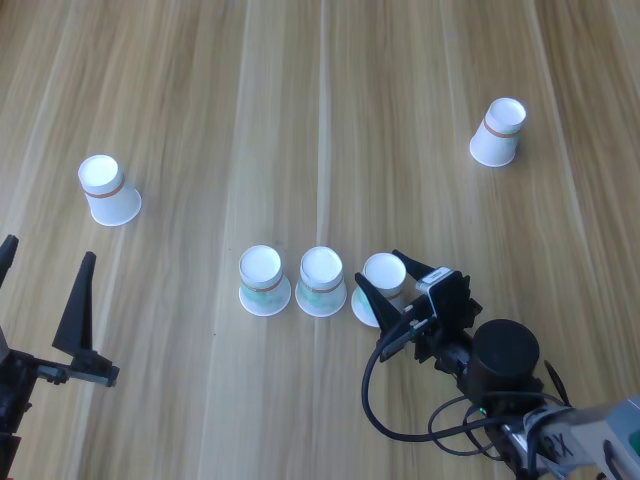}\hfill%
    \includegraphics[width=0.16\linewidth]{figure/real-task/cup/frame_001064.png}\hfill%
    \includegraphics[width=0.16\linewidth]{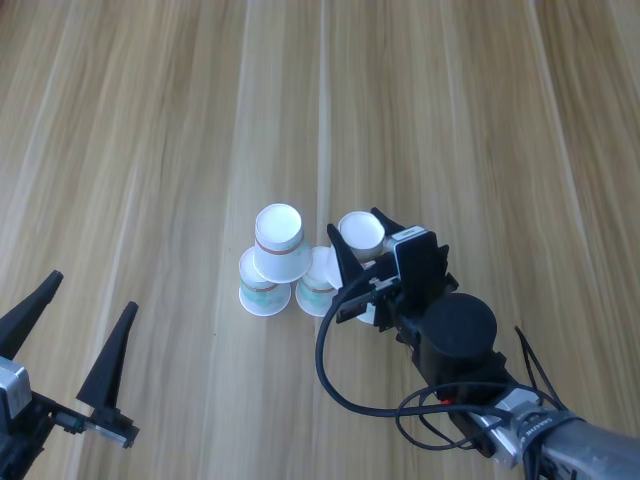}\hfill%
    \includegraphics[width=0.16\linewidth]{figure/real-task/cup/frame_001443.png}%
    \caption{\textbf{Paper-cup stacking.} The robot arranges three cups into a base and places two cups above them to form a two-tier stack. The six frames are ordered chronologically from left to right.}
    \label{fig:real_cup_stacking}
\end{figure}

\Needspace{0.34\textheight}
\subsection{Block stacking}
\label{sec:supp_block_stacking}

The goal is to stack two small blocks, red and yellow, on a larger blue base block (\Cref{fig:real_block_stacking}).
The blocks are initially separated on the tabletop.
The two small blocks are randomly positioned, while the large central base block position varies slightly.
In the illustrated sequence, the robot first picks up the red block, positions it over the blue block, and lowers it into contact before releasing it.
It then picks up the yellow block and places it on top of the red block to complete the three-block stack.
Each placement requires aligning the new block with the supporting block and withdrawing the gripper without disturbing the existing stack.
The desired final state is a stable tower containing all three blocks in the prescribed order.

\begin{figure}[H]
    \centering
    \includegraphics[width=0.16\linewidth]{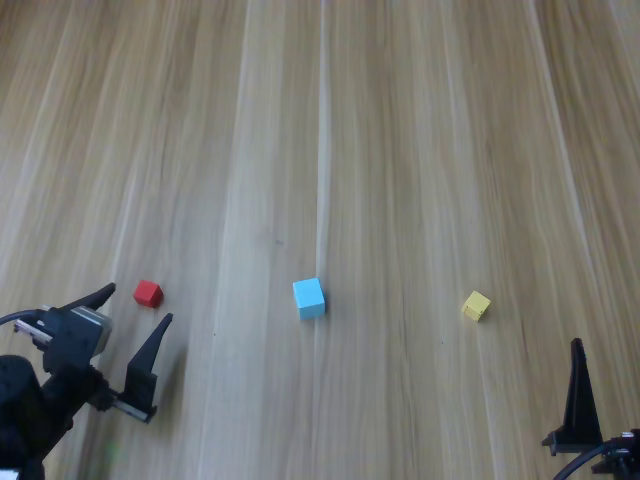}\hfill%
    \includegraphics[width=0.16\linewidth]{figure/real-task/cube/frame_000143.png}\hfill%
    \includegraphics[width=0.16\linewidth]{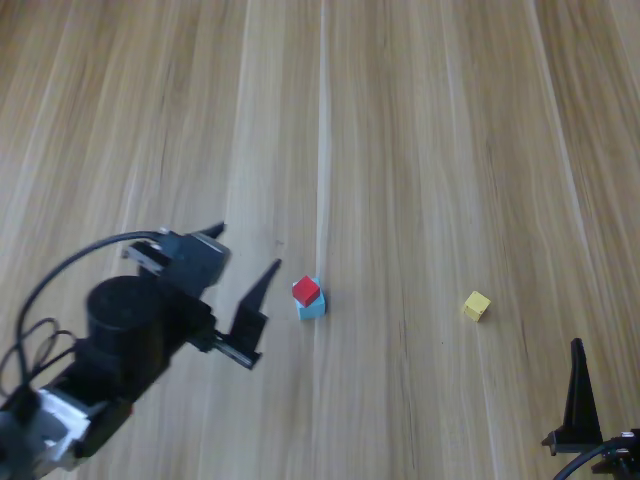}\hfill%
    \includegraphics[width=0.16\linewidth]{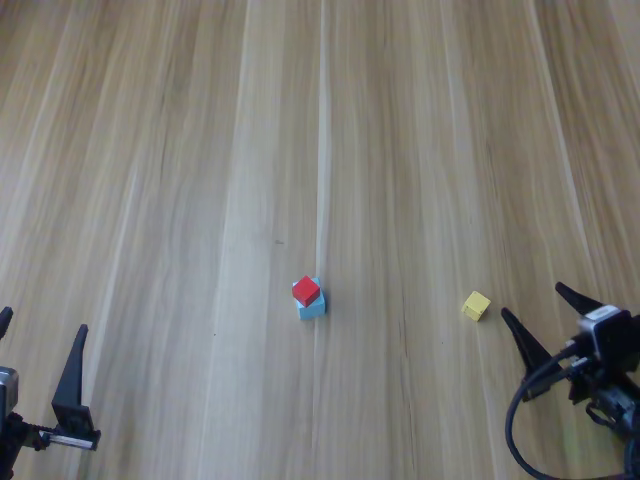}\hfill%
    \includegraphics[width=0.16\linewidth]{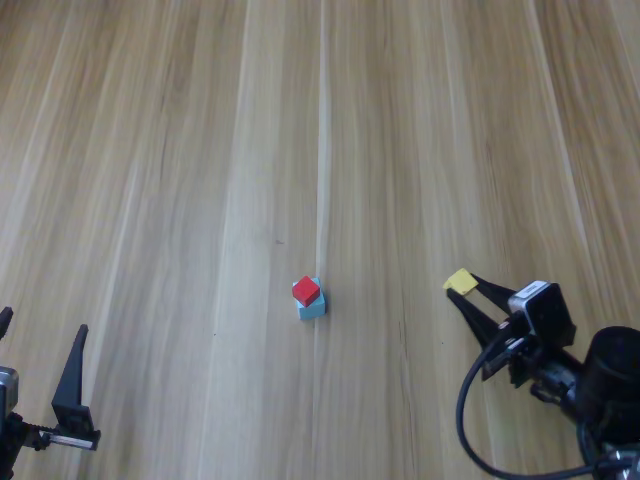}\hfill%
    \includegraphics[width=0.16\linewidth]{figure/real-task/cube/frame_000492.png}%
    \caption{\textbf{Block stacking.} The robot places the small red block on the larger blue base block and then adds the small yellow block. The six frames are ordered chronologically from left to right.}
    \label{fig:real_block_stacking}
\end{figure}

\FloatBarrier

\clearpage
\section{RoboTwin per-task results}
\label{sec:supp_robotwin_results}

\Cref{tab:robotwin_per_task} lists the clean and randomized success rates for all 50 RoboTwin tasks for \fdmvlm{} and \fdmvgm{}.
We compute each condition mean as the arithmetic mean of the 50 task success rates, and the overall mean as the arithmetic mean of the two condition means.
The resulting overall success rates are 81.20\% for \fdmvlm{} and 89.23\% for \fdmvgm{}, as reported in \Cref{tab:robotwin}.

\begin{table}[htbp]
    \caption{\textbf{Per-task success on RoboTwin 2.0.} Success rates are in percent. Each condition mean gives equal weight to all 50 tasks; the overall mean also weights clean and randomized conditions equally.}
    \label{tab:robotwin_per_task}
    \centering
    \scriptsize
    \renewcommand{\arraystretch}{0.95}
    \setlength{\tabcolsep}{7pt}
    \begin{tabular}{lrrrr}
    \toprule
    & \multicolumn{2}{c}{\fdmvlm{}} & \multicolumn{2}{c}{\fdmvgm{}} \\
    \cmidrule(lr){2-3}\cmidrule(lr){4-5}
    Task & Clean & Rand. & Clean & Rand. \\
    \midrule
    adjust\_bottle & 100 & 100 & 100 & 100 \\
    beat\_block\_hammer & 99 & 98 & 97 & 96 \\
    blocks\_ranking\_rgb & 91 & 90 & 100 & 99 \\
    blocks\_ranking\_size & 75 & 67 & 93 & 90 \\
    click\_alarmclock & 48 & 46 & 100 & 100 \\
    click\_bell & 94 & 90 & 100 & 100 \\
    dump\_bin\_bigbin & 99 & 98 & 93 & 92 \\
    grab\_roller & 100 & 100 & 100 & 100 \\
    handover\_block & 64 & 76 & 42 & 60 \\
    handover\_mic & 100 & 100 & 96 & 98 \\
    hanging\_mug & 45 & 48 & 55 & 62 \\
    lift\_pot & 11 & 6 & 98 & 94 \\
    move\_can\_pot & 55 & 65 & 91 & 99 \\
    move\_pillbottle\_pad & 85 & 85 & 98 & 99 \\
    move\_playingcard\_away & 99 & 96 & 99 & 100 \\
    move\_stapler\_pad & 47 & 52 & 65 & 76 \\
    open\_laptop & 98 & 100 & 99 & 100 \\
    open\_microwave & 72 & 56 & 18 & 27 \\
    pick\_diverse\_bottles & 70 & 69 & 77 & 85 \\
    pick\_dual\_bottles & 75 & 74 & 89 & 93 \\
    place\_a2b\_left & 92 & 95 & 94 & 94 \\
    place\_a2b\_right & 94 & 89 & 93 & 98 \\
    place\_bread\_basket & 87 & 81 & 92 & 97 \\
    place\_bread\_skillet & 78 & 86 & 90 & 93 \\
    place\_burger\_fries & 96 & 89 & 96 & 97 \\
    place\_can\_basket & 64 & 58 & 71 & 60 \\
    place\_cans\_plasticbox & 88 & 88 & 98 & 95 \\
    place\_container\_plate & 100 & 97 & 96 & 99 \\
    place\_dual\_shoes & 85 & 79 & 89 & 86 \\
    place\_empty\_cup & 99 & 97 & 99 & 100 \\
    place\_fan & 94 & 85 & 96 & 99 \\
    place\_mouse\_pad & 69 & 62 & 89 & 82 \\
    place\_object\_basket & 61 & 74 & 82 & 80 \\
    place\_object\_scale & 79 & 85 & 90 & 97 \\
    place\_object\_stand & 87 & 90 & 94 & 92 \\
    place\_phone\_stand & 97 & 95 & 99 & 100 \\
    place\_shoe & 95 & 97 & 96 & 97 \\
    press\_stapler & 100 & 99 & 93 & 93 \\
    put\_bottles\_dustbin & 62 & 60 & 92 & 73 \\
    put\_object\_cabinet & 80 & 80 & 89 & 89 \\
    rotate\_qrcode & 90 & 90 & 92 & 87 \\
    scan\_object & 71 & 62 & 93 & 82 \\
    shake\_bottle & 99 & 99 & 100 & 100 \\
    shake\_bottle\_horizontally & 100 & 100 & 100 & 100 \\
    stack\_blocks\_three & 87 & 82 & 95 & 96 \\
    stack\_blocks\_two & 99 & 97 & 99 & 100 \\
    stack\_bowls\_three & 74 & 83 & 73 & 82 \\
    stack\_bowls\_two & 91 & 95 & 97 & 93 \\
    stamp\_seal & 63 & 65 & 85 & 87 \\
    turn\_switch & 69 & 68 & 60 & 63 \\
    \midrule
    Condition mean & 81.54 & 80.86 & 88.84 & 89.62 \\
    Overall mean & \multicolumn{2}{c}{81.20} & \multicolumn{2}{c}{89.23} \\
    \bottomrule
    \end{tabular}
\end{table}

\FloatBarrier

\end{document}